\documentclass[a4paper,ngerman,12pt,chapterprefix=false,twoside,openright]{scrbook}
\usepackage[utf8]{inputenc}
\usepackage[ngerman, main=english]{babel}
\usepackage[headings]{fullpage}
\usepackage{helvet}

\usepackage{todonotes}
\usepackage{eurosym}
\usepackage{textcomp}
\usepackage{gensymb}
\usepackage{algorithm}
\usepackage{algpseudocode}

\graphicspath{{./src/pics/}}
\usepackage{pgfplots}
\pgfplotsset{width=0.83\linewidth, compat=1.9}
\usepackage{tikz} 
\usetikzlibrary{matrix,chains,positioning,decorations.pathreplacing,arrows}
\usepackage{verbatim}
\usepackage{subfigure}
\usepackage{float}
\usetikzlibrary{plotmarks,shapes,arrows.meta,decorations.markings,patterns,calc}

\usepackage{amsmath,amsfonts,amssymb}
\usepackage{mathtools}
\usepackage{bm}

\usepackage{tabularx}
\usepackage{booktabs}

\newenvironment{ptm}{\fontfamily{ptm}\fontsize{20}{24}\selectfont}{\par}
\newcolumntype{K}[1]{>{\centering\arraybackslash}b{#1}}

\makeindex

\usepackage{scrlayer-scrpage}
 \newcommand{\ifrtitle}{Observational and Interventional Causal Learning for Regret-Minimizing Control}
  \newcommand{\ifrauthor}{Christian Reiser}
   \newcommand{\ifrtype}{} 

 \newcommand{\ind}{\perp\!\!\!\!\perp} 
 \newcommand{\dep}{\not\perp\!\!\!\!\perp}

 \newcommand{\ATE}{\frac{1}{T} \sum_{t}X^y_t} 
  
 \newcommand{\bbR}{\bar{\bar{R}}}

\usepackage[pdfauthor={\ifrauthor},
          pdftitle={\ifrtitle},
          pdfsubject={\ifrtype},
          pdfkeywords={},
          pdfproducer={pdflatex},
          pdfcreator={Latex}]{hyperref}

\usepackage{blindtext}

\usepackage{units}
\usepackage[intoc]{nomencliFR_A4}

\newcommand{\nomenunit}[1]{\hskip-0em\parbox{4em}{$\left[ #1 \right]$}}

\makenomenclature
\usepackage[numbers]{natbib}
\newcommand{\chapterFoundations}{provides an overview over the theoretic foundations on which this thesis builds on, which are graphical terminology to understand causal graphs, structural causal models to describe data-generating processes, the theoretic background and description of a state of the art  causal discovery algorithm called LPCMCI, and interventional causal discovery with randomized control trails.}

\newcommand{\chapterRQOne}{describes how LPCMCI was extended to profit from interventional data.}

\newcommand{\chapterRQTwo}{explains with the help of numerical numerical simulations how the extended version of LPCMCI can be used for regret minimizing control.}

\newcommand{\chapterResults}{shows the results of the numerical experiments.}

\newcommand{\chapterDiscussion}{discusses the results, the limitations of the proposed method and gives ideas that can guide further research.}

\begin{document}

\frontmatter

\begin{titlepage}

\vspace*{2cm}
\begin{center}
\begin{ptm}
\begin{minipage}{\textwidth}
\begin{center}
\vspace{0.5cm}
 \textbf{\ifrtitle}
\vspace{0.5cm}
\end{center}
\end{minipage}

\vspace{4cm}
\textbf{\ifrauthor}\\
\vspace{1cm}
\textbf{\ifrtype}\\
\vspace{0.5cm}
\textbf{\the\year}
\\

\end{ptm}
\end{center}
\vfill
\begin{tabular}{K{2.7cm}K{13.0cm}}
	\begin{minipage}{\columnwidth}
	 \vspace{0.1cm}
	 \end{minipage}
	 & 
	\begin{minipage}{\columnwidth}
	\vfill
 \sffamily \large University of Stuttgart \\ \textbf{}\vspace{0.1cm}
	 \end{minipage}\\

	 \hline
 \end{tabular}

\end{titlepage}


\section*{Abstract}
Reconstructing the causal relationships behind the phenomena we observe is a fundamental challenge in all areas of science.
Experimentation, i.e., interventions through randomized control trials with large sample sizes, is considered the gold standard in discovering causal relationships.
However, they are also expensive and sometimes even ethically problematic.
In contrast, the emerging research area of observational causal discovery aims to detect causal relationships without experimentation. While observational causal discovery is cheaper and ethically less unproblematic, its conclusions are less robust and precise than randomized control trials.

The first part of this thesis explores the idea that both causal discovery methods can be combined to detect causal relationships reliably while reducing the number of experiments.
More precisely, the state-of-the-art observational causal discovery algorithm for time series capable of handling latent confounders and contemporaneous effects, called LPCMCI\cite{gerhardus_lpcmci_2020}, is extended to profit from casual constraints found through randomized control trials. 
I find that LPCMCI can be extended to use interventional data. Numerical results show that, given perfect interventional constraints, the reconstructed structural causal models (SCMs) of the extended LPCMCI allow $84.6\%$ of the time for the optimal prediction of the target variable. This result contrasts the baseline of $53.6\%$ when using the original LPCMCI algorithm that does not use interventional constraints. 
The implementation of interventional and observational causal discovery is modular, allowing causal constraints from other sources.

The second part of this thesis investigates the question of regret minimizing control by simultaneously learning a causal model and planning actions through the causal model.
The idea is that an agent with the objective of optimizing a measured variable first learns the system's mechanics through observational causal discovery. The agent then intervenes on the most promising variable with randomized values allowing for the exploitation and generation of new interventional data.
The agent then uses the interventional data to enhance the causal model further, allowing improved actions the next time.

The extended LPCMCI can be favorable compared to the original LPCMCI algorithm. 
The numerical results show that detecting and using interventional constraints leads to reconstructed SCMs that allow $60.9\%$ of the time for the optimal prediction of the target variable in contrast to the baseline of $53.6\%$ when using the original LPCMCI algorithm. 
Furthermore, the induced average regret decreases from $1.2$ when using the original LPCMCI algorithm to $1.0$ when using the extended LPCMCI algorithm with interventional discovery.

This project's source code is available \href{https://github.com/christianreiser/correlate/blob/master/causal_discovery/LPCMCI/plan.py}{online}.

\thispagestyle{empty}
\tableofcontents
\cleardoublepage

\addcontentsline{toc}{chapter}{List of Figures}
\listoffigures
\cleardoublepage

\addcontentsline{toc}{chapter}{Nomenclature}
\chapter*{Nomenclature}
\subsubsection*{Abbreviations}
\printnomenclature{A}
\subsubsection*{Latin Letters}
\printnomenclature{L}
\subsubsection*{Greek Letters}
\printnomenclature{G}
\subsubsection*{Indices}
\printnomenclature{I}
\subsubsection*{Mathematical Symbols}
\printnomenclature{S}

\cleardoublepage

\nomenclature[A]{DAG}{Directed acyclic graph}
\nomenclature[A]{CI}{Conditional independence}
\nomenclature[A]{MAG}{Maximal ancestral graph}
\nomenclature[A]{PAG}{Partial ancestral graph}
\nomenclature[A]{PDAG}{Partially directed acyclic graph}
\nomenclature[A]{SCM}{Structual causal model}
\nomenclature[A]{}{}
\nomenclature[A]{}{}
\nomenclature[A]{}{}
\nomenclature[A]{}{}

\nomenclature[L]{${f}$}{\nomenunit{-}Function or mechanism}
\nomenclature[L]{$N$}{\nomenunit{-}Number of measured variables}
\nomenclature[L]{$T$}{\nomenunit{-}Length of the timeseries}
\nomenclature[L]{$\Tilde{N}$}{\nomenunit{-}Number of data-generating variables including latents}
\nomenclature[L]{$\mathcal{N(\mu,\sigma)}$}{\nomenunit{-}Normal distibution with mean $\mu$ and standard deviation $\sigma$}
\nomenclature[L]{$\mathcal{U}$}{\nomenunit{-}Uniform distibution}
\nomenclature[L]{$\mathcal{M}$}{\nomenunit{-}A reconstructed structural causal model }
\nomenclature[L]{$\mathcal{M^*}$}{\nomenunit{-}The true data-generating structural causal model }
\nomenclature[L]{$\mathcal{M}_{do}$}{\nomenunit{-}A structural causal model with an intervention}
\nomenclature[L]{$\textbf{X}$}{\nomenunit{-}Multivariate timeseries of measurements}
\nomenclature[L]{$\textbf{X}^i$}{\nomenunit{-}Univariate timeseries of measurements}
\nomenclature[L]{$\textbf{V}$}{\nomenunit{-}Multivariate timeseries of the data-generator}
\nomenclature[L]{$\textbf{V}^i$}{\nomenunit{-}Univariate timeseries of the data-generator}
\nomenclature[L]{${X}^i_t$}{\nomenunit{-}Single measurement of variable $i$ at time $t$}
\nomenclature[L]{${V}^i_t$}{\nomenunit{-}Single generated value of variable $i$ at time $t$}
\nomenclature[L]{$p$}{\nomenunit{-}p-value}

\nomenclature[G]{$\tau$}{\nomenunit{-}Time lag between cause and effect}
\nomenclature[G]{$\tau_{max}$}{\nomenunit{-}Maximum time lag between cause and effect}
\nomenclature[G]{$\alpha$}{\nomenunit{-}Significance threshold for hypothesis tests}
\nomenclature[G]{$\eta$}{\nomenunit{-}Independently identically distributed noise}

\nomenclature[I]{$i$}{\nomenunit{-}Variable and node index}
\nomenclature[I]{$j$}{\nomenunit{-}Variable and node index}
\nomenclature[I]{$t$}{\nomenunit{-}Time index}

\nomenclature[S]{$\not\!\perp\!\!\!\perp$}{\nomenunit{-}Variable and node index}

\mainmatter

\chapter{Introduction}
\section{Learning through observation and interaction for decision making}
When we want to solve any problem, our first idea is to understand the mechanisms underlying the problem, especially the cause of the problem. We do that by learning through observing and interacting with the system. In our mind, we build a model of how the system could work. Then we try to solve the problem by taking the most promising action that our model allows. Whether the problem improves or not, the action produces new information, which we use to increase our understanding of cause and effect, about the consequence of the action. This process might help us to come up with an even better action for the next time.

Whether we are trying to find the best treatment for a disease or come up with the best policy to fight climate change, 
we first learn how the system behaves by itself, seek to influence it through our actions, and further learn from the outcomes of our actions.

\section{The use case of recommending actions to improve the mood of a person} 
Modern problems have become increasingly complex and data-intensive, and we struggle to understand the underlying mechanisms on our own.
The example case problem of this thesis is that people sometimes have days with generally elevated or depressed mood, are unsure about the reasons, and therefore do not know what to do about it.
A decrease in mood could have many potential causes. Examples are an allergy to pollen in the air, being sensitive to the weather, a food intolerance, impaired sleep quality, adverse effect of social media, too much exercise, or a lack of exercise.
In addition to a high number of variables to consider, finding the trigger is challenging because the time delay between the cause and effect can range from almost instantaneous to several days. 
For example, when a person is exposed to an allergen, it can take several days until the symptoms appear \cite{durieu_photoaggravated_2001}.

\section{Why not statistical models?}
\begin{figure}[htbp]
\begin{center}
\subfigure[Exercise influences mood.]{
    \begin{tikzpicture}[node distance={25mm}, thick, main/.style = {draw, circle}, hidden/.style = {draw=gray,dashed, circle}] 
        \node[main] (0) {mood}; 
        \node[main] (1) [left of=0]{exercise}; 
            \draw[->] (1) -- (0); 
    \end{tikzpicture}} \hspace{5mm}
\subfigure[Mood influences exercise.]{
    \begin{tikzpicture}[node distance={25mm}, thick, main/.style = {draw, circle}, hidden/.style = {draw=gray,dashed, circle}] 
        \node[main] (0) {mood}; 
        \node[main] (1) [left of=0]{exercise}; 
            \draw[<-] (1) -- (0); 
    \end{tikzpicture}} \hspace{5mm}
    
\subfigure[Weather acts as a confounder by influencing mood and exercise.]{
     \begin{tikzpicture}[node distance={25mm}, thick, main/.style = {draw, circle}, hidden/.style = {draw=gray,dashed, circle}] 
        \node[main] (0) {mood}; 
        \node[main] (4) [above right of=0] {weather}; 
        \node[main] (1) [below right of=4]{exercise}; 
            \draw[->] (4) -- (1); 
            \draw[->] (4) -- (0); 
    \end{tikzpicture}}\hspace{5mm}
\subfigure[All variables affect each other, which is one possible combination of the other cases.]{
     \begin{tikzpicture}[node distance={25mm}, thick, main/.style = {draw, circle}, hidden/.style = {draw=gray,dashed, circle}] 
        \node[main] (0) {mood}; 
        \node[main] (4) [above right of=0] {weather}; 
        \node[main] (1) [below right of=4]{exercise}; 
            \draw[<->] (4) -- (1); 
            \draw[<->] (4) -- (0); 
            \draw[<->] (1) -- (0); 
    \end{tikzpicture}}\hspace{5mm}
    
    \caption[Prediction vs. causality]{Possible explanations why exercise can be a good predictor of mood.}
    \label{fig:causal_directionalities}
\end{center}
\end{figure}
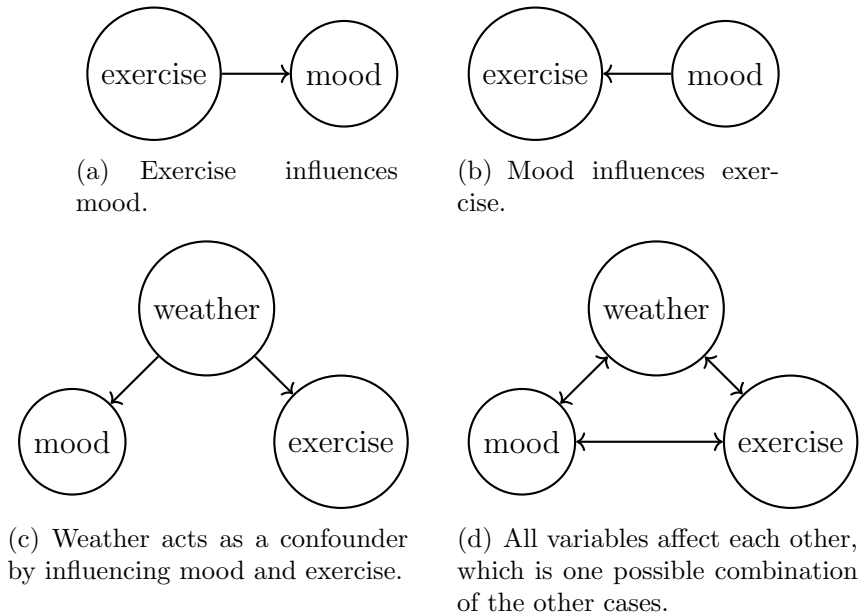

Typically supervised machine learning comes to mind when we want to solve data-intensive challenges. 
We track a person's mood and its potential influences that can be measured easily. 
A simple method would be to predict the mood from its potential influences via multiple linear regression and conclude that if a potential cause is important for prediction, it is related to mood swings.
The importance of a variable for prediction might tell us if or how strongly the variable is \textit{related} to mood. However, it does not necessarily tell us the \textit{causal direction} of the relationship, which is necessary to determine if an intervention on the variable would affect the person's mood.
In some instances, we know that some causal directions are not plausible. 
One example is that there could be a correlation between a weekend or a workday and a person's mood. 
In this case, it is obvious to us that a change in mood can not influence the day to be a weekend or workday.
Another example is when there is a time delay between two related variables. Then we know that the variable later in time cannot be the cause of the variable earlier in time.
A less obvious example is if we only see that a higher amount of exercise during one day is a good predictor of a good mood on the same day. This is a more challenging example as there is no measured time delay between both variables, and there are multiple possible explanations as illustrated in Figure \ref{fig:causal_directionalities}. 
Plot (a) shows that the amount of exercise could influence mood. In this case, a recommendation to change exercise habits could be helpful.
However, as shown in (b), mood could also influence how much a person exercises. For example, a person could exercise less because of a depressed mood. 
(c) depicts that there could also be a third variable, the weather, acting as a confounder, affecting how much someone exercises and their mood.
In (b) and (c), a recommendation to change exercise behavior will not be helpful.
Furthermore, there could also be any combination of the previous cases, e.g., all variables affect each other, as shown in (d).

More generally, the task of recommending interventions of what the person should do from passive observations of past behavior is categorized as prediction under intervention, for which purely statistical models are not suited \cite[section 1.4.1]{peters_elements_2018}. 
This would be an ill-posed task even with an infinite amount of passive observational data.
In contrast to statistical models, causal models like the causal graphs of Figure \ref{fig:causal_directionalities} can predict outcomes under interventions\cite[section 1.4.1]{peters_elements_2018}. 

\section{Causal discovery}
Now that we know that causal models can predict outcomes under interventions, the question is, how can we learn these causal models from data?
There is growing scientific progress in the field of \textit{discovering causal relationships from observational data}\cite{peters_elements_2018}. It can allow for gaining causal information by introducing assumptions on how the data was created.
However, there is often a tradeoff in making assumptions about the properties of the data-generating process. On the positive side, strong assumptions can lead to more precise results. On the negative side, the more assumptions we use, and the stronger the assumptions are, the higher the chances that an assumption may be wrong, leading to an increase in the probability of false results.
For example, an assumption we can make or not is that no confounder, like the weather in Figure \ref{fig:causal_directionalities}(c), exists but is unobserved. In other words, there are no latent confounders. Methods that do not make this assumption have more cases where they are unsure about the causal direction. Whereas methods that make the assumption are often more precise but lead more often to false results if there is an overlooked confounder \cite{gerhardus_lpcmci_2020}. In many real-world applications, the latter might often be the case as we often can not be sure if we really considered all relevant confounders.
A recent study shows that the results of observational causal discovery can provide helpful information, but the results are far from perfect\cite{reiser_causal_2022}. There, some predicted causal dependencies were false, and sometimes the causal direction could not be oriented.

\section{Randomized experiemnts}
\label{sec:rct}
Another way of determining causal relationships is through experimentation. One can precisely attribute any differences in the experiment's outcome to the intervention when using randomization to balance observed and unobserved characteristics. This form of experimentation is called a \textit{randomized control trial (RCT)}. In sufficiently large experiments, RTCs are robust and the gold standard in discovering causal effects. However, RCTs come at the cost of conducting experiments with many samples that must be carefully controlled by randomization, which is often tedious, expensive\cite{hariton_randomised_2018}, and sometimes ethically problematic.
Furthermore, there is the possibility that the experiment's outcome has negative consequences, for example, when a proposed medication has severe side effects. Another negative aspect is that RCT trials are mainly about understanding the effect of actions but not yet about taking the most promising action. For example, during the COVID-19 pandemic, some people who enrolled in RCTs for vaccines were not vaccinated as they received a placebo instead.

\section{Metrics}
To quantify how good actions are, I compute a unit called \textit{regret}, which is the difference between the outcomes of the actions taken and the theoretical outcomes if the optimal action would be taken every time.
Actions can be non-optimal either when a random action is non-optimal or when the causal information is incorrect or imprecise.



\section{The research question}
One can see that observational causal discovery and interventional causal discovery (e.g., RCTs) have different strengths and weaknesses. 
Interventional causal discovery is the more robust method to detect and quantify causal effects, but observational causal discovery is less expensive and more convenient. Furthermore, in contrast to observational discovery, RCTs can have downsides, as placebos or alternative treatments can be useless or suboptimal w.r.t. the outcome, and new interventions can have unknown harmful consequences.

One can see that observational causal discovery has a weakness, where RCT has its strength and vice versa. RCTs are robust and precise but also expensive, while observational causal discovery is more prone to errors and often imprecise but cheap and more convenient as no experiments are needed. 
This points to the question if both methods can be combined to profit from each other.
This leads to the two research questions of this thesis.
The first research question is if a state-of-the-art observational causal discovery algorithm for multivariate time series with latent confounders, called LPCMCI, can be extended to also profit from interventional data.
The second research question is if this extension of LPCMCI discovers causal models that lead to less regret when used for control compared to the original version.

\section{The structure of the thesis}
This thesis is structured in the following way.
Chapter \ref{ch:Foundations} \chapterFoundations{}
Chapter \ref{ch:RQOne} \chapterRQOne{}
Chapter \ref{ch:RQTwo} \chapterRQTwo{}
Chapter \ref{ch:results} \chapterResults{}
In Chapter \ref{ch:discussion} \chapterDiscussion{}

\chapter{Theoretic foundations and literature review}
\label{ch:Foundations}

\section{Graph terminology}
\label{sec:graph}

Our first research question requires causally modeling complex systems. It is often difficult to understand such models from equations and more intuitive when seeing representations of the causal system in the form of graphs. Therefore, this section explains the necessary terminology of graphs, how graphs can describe causal models, and how to describe causal models in equations.

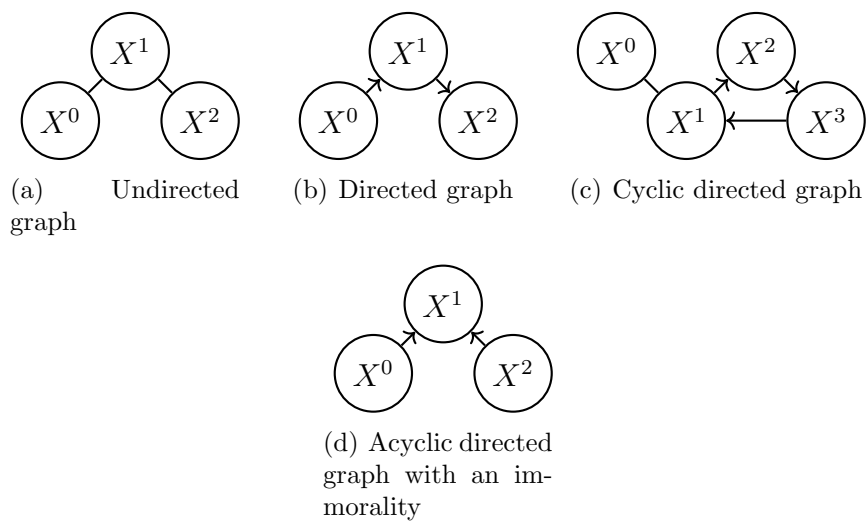
\begin{figure}[htbp]
\begin{center}

\subfigure[Undirected graph]{
     \begin{tikzpicture}[node distance={13mm}, thick, main/.style = {draw, circle}, hidden/.style = {draw=gray,dashed, circle}] 
        \node[main] (0) {$X^0$}; 
        \node[main] (4) [above right of=0] {$X^1$}; 
        \node[main] (1) [below right of=4]{$X^2$}; 
            \draw[-] (4) -- (1); 
            \draw[-] (4) -- (0); 
    \end{tikzpicture}}\hspace{5mm}
\subfigure[Directed graph]{
     \begin{tikzpicture}[node distance={13mm}, thick, main/.style = {draw, circle}, hidden/.style = {draw=gray,dashed, circle}] 
        \node[main] (0) {$X^0$}; 
        \node[main] (4) [above right of=0] {$X^1$}; 
        \node[main] (1) [below right of=4]{$X^2$}; 
            \draw[->] (4) -- (1); 
            \draw[<-] (4) -- (0); 
    \end{tikzpicture}}\hspace{5mm}
\subfigure[Cyclic directed graph]{
     \begin{tikzpicture}[node distance={13mm}, thick, main/.style = {draw, circle}, hidden/.style = {draw=gray,dashed, circle}] 
        \node[main] (0) {$X^1$}; 
        \node[main] (5) [above left of=0] {$X^0$}; 
        \node[main] (4) [above right of=0] {$X^2$}; 
        \node[main] (1) [below right of=4]{$X^3$}; 
            \draw[->] (4) -- (1); 
            \draw[<-] (4) -- (0); 
            \draw[-] (5) -- (0); 
            \draw[->] (1) -- (0); 
    \end{tikzpicture}}\hspace{5mm}
    
    \subfigure[Acyclic directed graph with an immorality]{
     \begin{tikzpicture}[node distance={13mm}, thick, main/.style = {draw, circle}, hidden/.style = {draw=gray,dashed, circle}] 
        \node[main] (0) {$X^0$}; 
        \node[main] (1) [above right of=0] {$X^1$}; 
        \node[main] (2) [below right of=1]{$X^2$}; 
            \draw[->] (0) -- (1); 
            \draw[->] (2) -- (1); 
    \end{tikzpicture}}\hspace{5mm}
    
    \caption[Graph terminology]{Types of graphs}
    \label{fig:graph}
\end{center}
\end{figure}

Consider Figure \ref{fig:graph} (a).
It shows a \textit{graph}, that is, a set of \textit{nodes} and \textit{edges} connecting nodes.
Nodes represent variables, and edges represent statistical dependencies between the pairs of variables.
Two nodes connected by an edge are \textit{adjacent}.
A graph is \textit{fully connected} when all pairs of nodes are adjacent.
A sequence of adjacent nodes, like $X^0-X^1-X^2$, is called a \textit{path} \cite[]{peters_elements_2018}.
(a) is called an \textit{undirected graph}, as its edges have no edgemarks.
In contrast, (b) is a \textit{directed graph} as its edges have edgemarks.
A directed edge goes out from a \textit{parent} node and points to the \textit{child} node. We will also denote the parent of $x^1$ as $pa(X^1)$.
A \textit{directed edge} represents a causal mechanism where the parent is a cause of the child. This means a change in the value of the parent can lead to a change in the child's value.
Graph (b) shows a \textit{directed path} $X^0\rightarrow X^1 \rightarrow X^2$, which is a path that consists of directed edges that all point in the same direction.
$X^0$ is an \textit{ancestor} of $X^2$ (also denoted as $X^0 \in an(X^2)$\cite{gerhardus_characterization_2021}), and $X^2$ is a \textit{descendant} of $X^0$ ($X^2 \in de(X^0)$) because there is a directed path that starts at node $X^0$ and ends at node $X^2$. By the same reasoning, $X^1 \in an(X^2)$, and $X^2 \in de(X^1)$. 

Figure (c) shows a \textit{partially directed graph} with a\textit{directed cycle}. A partially directed graph without cycles is known \textit{partially directed acyclic graph (PDAG)}. A PDAG where all edges are directed is a \textit{directed acyclic graph (DAG)}. A sample DAG is illustrated in Figure (d).
DAGs are introduced here because they are used to visualize data-generating processes intuitively. 
The goal of causal discovery is to reconstruct the data-generating process, ideally finding the DAG representing the data generator.

\label{sec:implLatent}

\begin{figure}[htbp]
    \centering
             \subfigure[Example of a DAG that represents a structural causal model with two hidden variables.]{
   \begin{tikzpicture}[node distance={15mm}, thick, main/.style = {draw, circle}, hidden/.style = {draw=gray,dashed, circle}] 
\node[main] (0) {$X^0$}; 
\node[main] (1) [right of=0]{$X^1$}; 
\node[hidden] (2) [above right of=1] {$H^0$}; 
\node[hidden] (3) [below right of=1] {$H^1$}; 
\node[main] (4) [above right of=3] {$X^2$}; 
\draw[->] (0) -- (1); 
\draw[<-] (1) -- (2); 
\draw[<-] (1) -- (3);
\draw [->](2) -- (4); 
\draw [->](3) -- (4); 
\end{tikzpicture}}\hspace{5mm}
    \subfigure[The MAG over the observed variables of the DAG.]{
   \begin{tikzpicture}[node distance={15mm}, thick, main/.style = {draw, circle}, hidden/.style = {draw=gray,dashed, circle}] 
\node[main] (0) {$X^0$}; 
\node[main] (1) [right of=0]{$X^1$}; 
\node[main] (4) [above right of=3] {$X^2$}; 
    \draw[->] (0) -- (1); 
    \draw[<->] (1) -- (4); 
\end{tikzpicture}}\hspace{5mm}
    \subfigure[Oracle PAG]{
   \begin{tikzpicture}[node distance={15mm}, thick, main/.style = {draw, circle}, hidden/.style = {draw=gray,dashed, circle}, o/.style={
        shorten >=#1,
        decoration={
            markings,
            mark={
                at position 1
                with {
                    \draw circle [xshift=-#1,radius=#1];
                }
            }
        },
        postaction=decorate
    },
    o/.default=2pt] 
\node[main] (0) {$X^0$}; 
\node[main] (1) [right of=0]{$X^1$}; 
\node[main] (4) [above right of=3] {$X^2$}; 
    \draw[->] (0) -- (1); 
    \draw[o] (1) -- (0); 
    \draw[o] (1) -- (4); 
    \draw[<-] (1) -- (4); 
\end{tikzpicture}
    }
    \caption[Three types of graphs: DAG, MAG, PAG]{A comparison of three graphs representing information about a structural causal model with latent variables.}
        \label{fig:DAGMAG}
\end{figure}
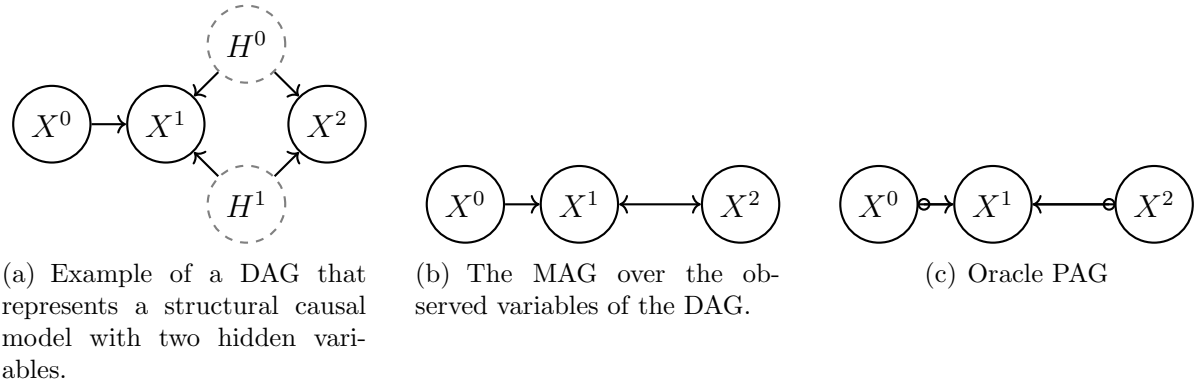

In this thesis, we will assume that some variables of data generators might not be measured. We call such variables \textit{latent} or \textit{hidden} variables.
When two variables have a common cause, the common cause is also called a \textit{confounder} of the two variables\cite[Section 9.1]{peters_elements_2018}.
Figure \ref{fig:DAGMAG}(a) shows a DAG with two \textit{latent confounders}, $H^0$ and $H^1$. 
In causal discovery, we ideally would like to reconstruct DAGs with latent confounders from the measurements of its observed variables. 
While the causal discovery algorithm of this thesis can often detect that measured variables are latently confounded, it can not identify the number of latent variables. This inability leads to the problem that there are an infinite number of DAG candidates.
Rather than trying to learn the DAG with an unknown number of latents, 
a feasible goal is to learn as many causal features as possible. 
Ideally, we learn a unique graph 
over the observed variables
that represents the causal constraints entailed by the original DAG\cite{zhang_completeness_2008}.

Therefore, we need to introduce a graph type capable of representing each DAG with latent variables by marginalizing over the observed variables\cite[Section 9.4]{peters_elements_2018}.
An example shows Figure \ref{fig:DAGMAG}, where in Figure (a) latent confounders $H^0$ and $H^1$ of the DAG are influencing the observed variables $X^1$ and $ X^2$, while graph (b) represents this constellation by a MAG (a type of graph introduced later) over the observed variables and indicating an arbitrary number of latents affecting $X^1, X^2$ with a \textit{bi-directional edge} $\leftrightarrow$.
Before we can define this MAG, we first need the following definitions.

\textit{Mixed graphs} are partially directed graphs with additional bi-directed edges.
A mixed graph is an \textit{ancestral graph} if the following two conditions hold. First, there are no directed cycles. Second, if there is a bidirected edge between two variables $X^0\leftrightarrow X^1$, then there is no directed path from $X^0$ to $X^1$, or from $X^1$ to $X^0$\cite{tian_generating_2012}.
A node $X^0$ in an ancestral graph is a \textit{collider} if two directed edges point to $X^0$\cite{zhang_causal_2008}. For example, node $X^1$ of Figure \ref{fig:graph} (d) is a collider.
In an ancestral graph, a path between two nodes $X^0$ and $X^1$ is \textit{m-connecting} relative to a possibly empty set of nodes $\mathbf{S}$, with $X^0, X^1 \notin \mathbf{S}$ if the following two conditions hold:
\begin{enumerate}
\item every node on the path that is not a collider is not in $\mathbf{S}$,
\item every collider on the path is in $\mathbf{S}$ or an ancestor of a node in $\mathbf{S}$ \cite{zhang_causal_2008}.
\end{enumerate}
The critical thing about m-connection is that m-connection depends on whether there is a collider on the path and if the collider or a descendant of the collider is in $\mathbf{S}$.
If one of the above conditions does not hold, then the path is not m-connecting and $\mathbf{S}$ \textit{blocks} the path.
Nodes $X^0$ and $X^1$ are \textit{m-separated} by $\mathbf{S}$ if there is no m-connecting path between $X^0$ and $X^1$, meaning $\mathbf{S}$ blocks all paths between $X^0$ and $X^1$.
Two disjoint subgraphs of nodes $\mathbf{X^0}$ and $\mathbf{X^1}$ are m-separated by $\mathbf{S}$ if every node in $\mathbf{X^0}$ is m-separated from every node in $\mathbf{X^1}$ by $\mathbf{S}$\cite{zhang_causal_2008}. 

m-separation in graphs is essential because a data generator with certain m-separations in its ancestral graph produces a distribution with conditional independence where there are m-separations in the graph.
More specifically, if sets $\mathbf{X^0}$ and $\mathbf{X^1}$ are m-separated by $\mathbf{S}$ in an ancestral graph G, then $\mathbf{X^0}$ is independent of $\mathbf{X^1}$ conditional on $\mathbf{S}$ in the underlying distribution P.
Formally that is:
$(\mathbf{X^0} \!\perp\!\!\!\perp \mathbf{X^1} \mid \mathbf{S})_{G} \Longrightarrow(\mathbf{X^0} \!\perp\!\!\!\perp \mathbf{X^1} \mid \mathbf{S})_{P}$ whenever $G$ and $P$ are compatible. 
Conversely, if $\mathbf{X^0}$ and $\mathbf{X^1}$ are not m-separated by $\mathbf{S}$ in G, then $\mathbf{X^0}$ and $\mathbf{X^1}$ are dependent conditional on $\mathbf{S}$ in the underlying distribution P.
Or formally, if $(\mathbf{X^0} \!\perp\!\!\!\perp \mathbf{X^1} \mid \mathbf{S})_{P}$ holds in all distributions compatible with $G$, it follows that $(\mathbf{X^0} \!\perp\!\!\!\perp \mathbf{X^1} \mid \mathbf{S})_{G}$\cite{richardson_ancestral_2002}\cite[Theorem 1.2.3]{pearl_causality_2000}.
I summarize what we learned, as their consequences are essential for constraint-based causal discovery. We know whether the distributions of two variables on a path are conditionally independent if we know which nodes of the path are colliders and which are in the conditioning set $\mathbf{S}$.
As a motivating outlook, we go in the other direction during causal discovery: If we find that in a triplet of variables $X^0-X^1-X^2$, and there is an independency between $X^0$ and $X^2$ when conditioning on $X^1$, then, with the help of additional assumptions introduced later, we know that $X^1$ is a collider and can direct the edges to $X^0\rightarrow X^1\leftarrow X^2$.

An ancestral graph is a \textit{maximal ancestral graph (MAG)} if for every pair of non-adjacent nodes $X^i, X^j$ in the graph there exists a set $\mathbf{S}$, with $X^i, X^j \notin \mathbf{S}$, such that $X^i, X^j$ are m-separated conditional on \textbf{S}\cite{zhang_completeness_2008}. This means that nodes in a MAG are connected by an edge if and only if they cannot be m-separated by a subset of observed variables. Consequently, when there is no edge between two variables in a MAG, then they are conditionally independent.
MAGs are a superclass of DAGs, as a DAG is also a MAG without bi-directed edges. However, MAGs are less precise and have the following differences in semantics. 
In contrast to a DAG, a directed edge's tail (-) does not indicate parentship anymore. Instead, it indicates only ancestry, which is less precise. For example, in $X^0 \rightarrow X^1$, while $X^0$ would be a parent of $X^1$ in a DAG, in a MAG, we have more uncertainty and only know $X^0$ is an ancestor of $X^1$. Furthermore, the arrowhead ($<$ or $>$) of a directed edge in a MAG, like the one pointing on $X^1$, does not necessarily point to the child or descendant but only specifies that $X^1$ is not an ancestor of $X^0$ (also denoted as $X^1 \notin an(X^0)$). 
In addition to the possibility that $X^1$ is a descendant of $X^0$, there is the possibility that the dependency might be due to confounding.
Since an arrowhead in a MAG declares the variable next to it as a non-ancestor, bidirectional edges ($\leftrightarrow$) exclude both variables from being the ancestor of the other variable and thus represent only the possibility of latent confounding between the two \cite{richardson_ancestral_2002}.
As an example, Figure \ref{fig:DAGMAG} shows how a MAG (Subfigure (b)) represents the information of the observed variables of a DAG that has latent confounders (Subfigure (a)).

The causal discovery algorithm we use in this thesis relies mainly on exploiting m-separations to gain causal information.
While m-separations in a MAG uniquely define which nodes are adjacent, the edgemarks of the adjacencies may be different \cite{zhang_completeness_2008}.
MAGs with the same m-separations, but different edgemarks are comprised in a \textit{Markov equivalence class}\cite{richardson_ancestral_2002}.
This uncertainty means that if no other constraints are available, our causal discovery algorithm might not return one MAG but the information of all MAGs within its Markov equivalence class.
Table \ref{tab:equiq} shows examples of Markov equivalence for cases without latent confounding. 
\begin{table}[htbp] 
\begin{center}
    \caption[Markov equivalence classes]{Chains of both directions and forks cannot be differentiated via conditional independence (CI). However, they can be differentiated from colliders. $X\!\perp\!\!\!\perp Y$ and $X\!\perp\!\!\!\perp Y|Z$ denotes independence and conditional independence between the random variables X and Y, respectively.}
\begin{tabular}{@{}l|lll|l@{}}
\toprule
Name         & Chain   & Chain   & Fork    & Collider \\ \midrule
    Graph & $X\rightarrow Z \rightarrow Y$ & $X\leftarrow Z \leftarrow Y$ & $X\leftarrow Z \rightarrow Y$ & $X\rightarrow Z \leftarrow Y$ \\
    Independence & $X\not\!\perp\!\!\!\perp Y$    & $X\not\!\perp\!\!\!\perp Y$    & $X\not\!\perp\!\!\!\perp Y$   & $ \mathbf{X\!\perp\!\!\!\perp Y}  $       \\
    CI           & $X\!\perp\!\!\!\perp Y|Z$   & $X\!\perp\!\!\!\perp Y|Z$   & $X\!\perp\!\!\!\perp Y|Z$  & $\mathbf{X\not\!\perp\!\!\!\perp Y|Z}$      \\
    Equivalence & \multicolumn{3}{l|}{one Markov equivalence class} & other Markov equivalence class \\ \bottomrule
\end{tabular}
        \label{tab:equiq}
\end{center}
\end{table}

The information of one Markov equivalence class can be summarized in one partial ancestral graph.
\textit{A partial ancestral graph (PAG)} is a partially mixed graph for a Markov equivalence class that
\begin{enumerate}
\item has the same adjacencies as any MAG in the Markov equivalence class, and
\item has unambiguous arrowhead or tail edgemarks ($<,>,-$) at locations if and only if all MAGs in the Markov equivalence class have the same edgemarks at that locations\cite{bejos_maximal_2020}.
\end{enumerate} 
PAGs have the same semantics as MAGs except that a PAG can also represent uncertainty whether a link has an arrowhead or tail with an ambiguous circle-head ($\circ$). Figure \ref{fig:DAGMAG}(c) shows an example of a PAG with circle-head edgemarks. The following is an example of the functionality of circle-head edgemarks. The link $X^2 \circ$$\rightarrow X^3$ says that $X^3$ is not an ancestor of $X^2$, but we do not know if $X^2$ is an ancestor of $X^3$ or not. 
This case leaves us with three possibilities. First, there could be a directed path from $X^2$ to $X^3$; second, the dependency could only be due to latent confounding; third, it could be a combination of both.

For more efficient communication, I introduce the edgemark placeholder $\star$ to indicate that it could be any of the PAG edgemarks $(<,>,-,\circ)$.
To clarify, I repeat that a PAG can only contain the edgemarks $<,>,-,\circ$, but the star $\star$ is never written into a PAG, and we use the star only to indicate that a property or a function applies to any of the $<,>,-,\circ$ PAG edgemarks.
For example, when saying that we search for a $X^0\circ$$-$$\star X^1$ constellation in a PAG, it means that we are searching for $X^0\circ$$-$$\circ X^1$, $X^0\circ$$\rightarrow X^1$, or $X^0\circ$$-$$- X^1$ in the PAG. 

Using the star edgemark, we define unshielded triples, which are triplets of the form $X^0\star -\star X^1\star -\star X^2$ that have no adjacency between $X^0$ and $X^2$. Unshielded triplets will be necessary for edgemark orientation \cite{gerhardus_lpcmci_2020}.

We defined all graphical terminology needed in this thesis, including that a directed edge represents a causal mechanism. The following section explains causal mechanisms and how to write them as equations.

\section{Structural causal models}
\label{sec:scm}
The equations of mechanisms in a system can be written in the form of a \textit{structural causal model (SCM)}:
\begin{equation}
    X^i:=f_i(pa(X^i, \eta^i),
    \label{eq:se}
\end{equation}
with the mechanisms (or functions) $f_i$ mapping parental variables and noises $pa(X^i, \eta^i)$ to the children $X^i$.
Note the "$:=$" instead of the "$=$" symbol, as the equation is not symmetric but has a direction just like a directed edge.
Equation \ref{eq:se} can generally be nonlinear\cite[Section 1.4.1]{pearl_causality_2000}; however, in the numerical experiments of this thesis, I only use linear functions for $f_i$.

One of the goals of this thesis is to model how the state of systems, such as a person's body and its environment, changes over time.
Consider a causal system as a multivariate state generator and assume the variables follow a vector-auto-regressive process\footnote{Auto-regressive process is a model where the output variable depends only on its own lagged values and on a stochastic noise term. A \textit{vector-}auto-regressive process extends the auto-regressive model to multiple dimensions. As explained later, the output of our vector-auto-regressive process additionally depends on its current values.}.
We can describe this process by adding a time dimension to the structural causal model (SCM) of equation \ref{eq:se}:
\begin{equation}
\label{eq:scm}
V_{t}^{j}:=f_{j}\left(pa\left(V_{t}^{j}\right), \eta_{t}^{j}\right) \quad \text { with } j=0, \ldots, \tilde{N}-1,
\end{equation}
 and timestep $t$.
The SCM can describe the state of a system for each timestep by generating a multivariate time series $\mathbf{V}^{j}=\left(V_{t}^{j}, V_{t-1}^{j}, \ldots\right)$ for variable indices $j=1, \ldots, \tilde{N}$.
The mechanisms $f_i$ depend on a set of causal parents $pa\left(V_{t}^{j}\right) \subseteq\left(\mathbf{V}_{t}, \mathbf{V}_{t-1}, \ldots, \mathbf{V}_{t-p_{t s}}\right)$ and jointly independent noise variables $\eta_{t}^{j}$. 
The noise terms represent other unknown causal influences that are assumed to be independent of each other.
$p_{t s}$ represents the maximal time lag between cause and effect. \cite{gerhardus_lpcmci_2020}

A SCM can be \textit{causally stationary}, which means that mechanisms between all a pairs of variables $\left(V_{t-\tau}^{i}, V_{t^{}}^{j}\right)$ with the time lag $\tau \geq 0$ are the same as for all time-shifted pairs $\left(V_{t^{\prime}-\tau}^{i}, V_{t^{\prime}}^{j}\right)$\cite{runge_causal_2018}.

\begin{figure}[htbp]
    \centering
    \subfigure[Time-series graph]{
   \begin{tikzpicture}[node distance={15mm}, thick, main/.style = {draw, circle}, hidden/.style = {draw=gray,dashed, circle}] 
\node[main] (0) {$X^0_0$}; 
\node[main] (10) [above of=0]{$X^1_0$}; 

\node[main] (1) [right of=0]{$X^0_1$}; 
\node[main] (11) [above of=1]{$X^1_1$}; 

\node[main] (2) [right of=1]{$X^0_2$}; 
\node[main] (12) [above of=2]{$X^1_2$}; 

\node[main] (3) [right of=2]{$X^0_3$}; 
\node[main] (13) [above of=3]{$X^1_3$}; 

    \draw[->] (0) -- (10); 
    \draw[->] (1) -- (11); 
    \draw[->] (2) -- (12); 
    \draw[->] (3) -- (13); 
\end{tikzpicture}}\hspace{15mm}
    \subfigure[Process graph]{
   \begin{tikzpicture}[node distance={15mm}, thick, main/.style = {draw, circle}, hidden/.style = {draw=gray,dashed, circle}, o/.style={
        shorten >=#1,
        decoration={
            markings,
            mark={
                at position 1
                with {
                    \draw circle [xshift=-#1,radius=#1];
                }
            }
        },
        postaction=decorate
    },
    o/.default=2pt] 
\node[main] (0) {$X^0$}; 
\node[main] (10) [above of=0]{$X^1$}; 
    \draw[->] (0) -- (10); 

\end{tikzpicture}
    }
    \caption[Time series graph and process graph]{The same graph is represented once as a time-series graph and once as a process graph.}
        \label{fig:process-graph}
\end{figure}
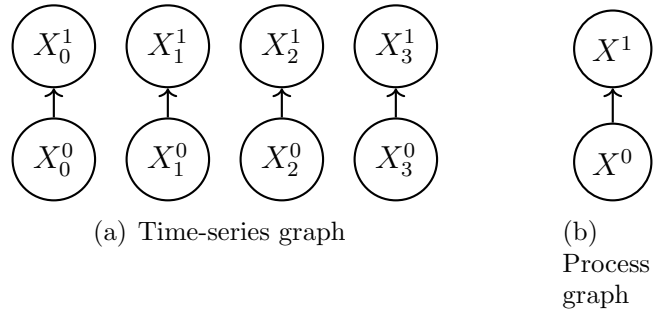

The SCM can be represented as a graph with an infinite repeating structure along the time axis. This representation is also known as a \textit{time-series graph}.
When the SCM is causally stationary, its representation as a time-series graph has no new information along the time axis. 
Therefore it is often more convenient to collapse the time-series graph along the time axis, which yields a \textit{process graph}. Figure \ref{fig:process-graph} illustrates a time-series graph with its corresponding process graph.

In this thesis, we translate MAGs and their mechanisms to SCMs without noise.
This process is possible as a MAG contains all parental information $pa(V^j_t)$, the mechanisms determine all $f_j$, and all noise variables $\eta_t^j=0$.

\section{Observational causal discovery with LPCMCI}
\label{sec:lpcmci}
\subsection{Overview}
LPCMCI is an observational causal discovery algorithm for multivariate time-series \cite{gerhardus_lpcmci_2020}. 
The strengths of LPCMCI are that it does not assume the absence of latent confounders or contemporaneous links, and it excels when the time series is highly autocorrelated.

Like all current state-of-the-art constraint-based causal discovery methods that allow for latent confounders and contemporaneous links, LPCMCI often cannot find the one MAG that describes the data-generating SCM, even with an unlimited sample size.
Instead, it aims to identify properties in the observational distribution which, together with assumptions, limit the number of possible MAGs to a Markov equivalence class represented by a PAG.


\subsection{Assumptions}
\label{sec:assumptions}
To causally reconstruct the data-generator described in Section \ref{sec:data-gen} from observational data, one needs to make assumptions about the data \cite[Chapter 2]{peters_elements_2018}. More specifically, these are assumptions about the original data generator and how the data was measured or preprocessed (e.g., data aggregation).

The first assumption we use is called \textit{faithfulness}. It means we are faithful that conditional independencies in the observed distribution
are due to m-separations in the original causal graph \cite{gerhardus_lpcmci_2020}.
Unfortunately, this assumption can be violated when conditional independencies in the observed distribution happen due to mere chance. 
However, the probability of violation becomes increasingly unlikely with larger sample sizes.
The benefit of faithfulness is that when we detect conditional independence in the observed distribution, we can constrain the causal graph that we want to estimate via an m-separation.

A fundamental property of LPCMCI is that it is unnecessary to assume that there are no \textit{latent variables} in the data-generating SCM. I use the words latent variables, latents, hidden variables, and unmeasured variables interchangeably. The existence of latents means that the measured time-series consists only of a subset $\mathbf{X}=\left\{\mathbf{X}^{1}, \ldots, \mathbf{X}^{N}\right\} \subseteq \mathbf{V}=\left\{\mathbf{V}^{1}, \mathbf{V}^{2}, \ldots\right\}$ of the generated time series with $N \leq \tilde{N}$.

The second assumption LPCMCI relies on is the absence of \textit{selection variables}, meaning that there are no variables determining which measurements are included or excluded from the data sample.
In most long-term real-world applications, some data points are expected to be missing. When using LPCMCI, it is essential to check that the data is missing at random and that no variable causes these missing data. A counterexample in my application is when a person does not track data when in a bad mood. Then there would be the selection variable mood, leading to selection bias in the data. In practice, one has to be careful if there is missing data because often, it only seems like data is missing at random. Still, there is a non-obvious reason and, thus, a selection variable. 

The third assumption LPCMCI uses is called \textit{causal stationarity} of the data generator. 
In Section \ref{sec:scm}, we learned that SCMs are causally stationary when the mechanisms do not change over time. 
This property simplifies the causal discovery problem by reusing a momentary constraint found at one point in the time series and setting it to all other time-shifted pairs.



Causes happen before effects ($\tau>0$). Therefore, it is possible to distinguish cause from effect if one can detect which variable changes first in time.
However, the measurement frequency can be too low. More precisely, the timesteps of the measured time series can be so large that measurements of the cause have identical timestamps as the effect values. 
Consequently, there seem to be \textit{contemporaneous effects} in the measured data ($\tau=0$) that remove the possibility of distinguishing cause from effect via time ordering.
For example, music can change a person's mood within minutes \cite{xiao_what_2008}. However, the mood is tracked often only once a day as it is more convenient than more often. Then the data shows that the time difference between changes in music and mood is $0$ days and therefore seems contemporaneous.
LPCMCI allows the possibility of most contemporaneous effects. 
The only exception is that LPCMCI assumes \textit{acyclicity}, which means there are no contemporaneous cycles \cite{peters_causal_2013}.
An example of a contemporaneous cycle is when the music causes a change in the mood of a person, and then the person's mood again causes the person to change the music.

\subsection{Causal discovery}
Most constraint-based observational causal discovery methods rely on the same main ideas but have differences in increasing detection power. First, I will explain the basic ideas and then the differences concerning LPCMCI. \
In short, the typical main idea of constraint-based causal discovery methods is first to start from a fully connected graph over the measured variables and apply time constraints. Then in the first phase, the edges will be removed, and in the second phase, the remaining edges will be oriented.

I now describe this process in more detail.
A working memory graph $\mathcal{C}(\mathcal{G})$ is initialized with $X_{t-\tau}^{i}  \circ$$-$$\circ X_{t}^{j} $ 
for $\tau=0$ 
and with $X_{t-\tau}^{i} \circ$$\rightarrow X_{t}^{j}$ 
for $0<\tau \leq \tau_{\max }$
which is a fully connected graph except for the time-constrained lagged links.
To remove edges, the methods test for independencies of adjacent pairs of variables $\left({{X}}^i_{t-\tau}\star- \star{{X}}^j_{t}\right)$ when conditioning on a set of other variables $\mathbf{S}$. The conditioning set $\mathbf{S}$ can also be empty.
If a pair of variables are conditionally independent, the corresponding nodes in the graph are m-separated, and the edge is removed. Furthermore, the variables in the conditioning set $\mathbf{S}$ are remembered as they will be used in the second phase.
    Remember, at the beginning of the algorithm, the graph is fully connected (except for time ordering constraints), and therefore most of its subgraphs are also fully connected. When edge removal during the first phase removes edges of complete subgraphs with three variables, these triplets become unshielded. Unshielded triplets, together with the information of the conditioning set, are essential for edge orientation in phase two.
    \begin{figure}[htbp]
    \centering
    \subfigure[Complete subgraph]{
   \begin{tikzpicture}[node distance={15mm}, thick, main/.style = {draw, circle}, hidden/.style = {draw=gray,dashed, circle}] 
\node[main] (0) {$X^0$}; 
\node[main] (1) [above right of=0]{$X^1$}; 
\node[main] (2) [below right of=1]{$X^2$}; 

    \draw[-] (0) -- (1); 
    \draw[-] (1) -- (2); 
    \draw[-] (2) -- (0); 
\end{tikzpicture}}\hspace{15mm}
    \subfigure[Unshielded triplet]{
   \begin{tikzpicture}[node distance={15mm}, thick, main/.style = {draw, circle}, hidden/.style = {draw=gray,dashed, circle}, o/.style={
        shorten >=#1,
        decoration={
            markings,
            mark={
                at position 1
                with {
                    \draw circle [xshift=-#1,radius=#1];
                }
            }
        },
        postaction=decorate
    },
    o/.default=2pt] 
\node[main] (0) {$X^0$}; 
\node[main] (1) [above right of=0]{$X^1$}; 
\node[main] (2) [below right of=1]{$X^2$}; 

    \draw[-] (0) -- (1); 
    \draw[-] (2) -- (0); 
\end{tikzpicture}}\hspace{15mm}

    \subfigure[Collider]{
   \begin{tikzpicture}[node distance={15mm}, thick, main/.style = {draw, circle}, hidden/.style = {draw=gray,dashed, circle}, o/.style={
        shorten >=#1,
        decoration={
            markings,
            mark={
                at position 1
                with {
                    \draw circle [xshift=-#1,radius=#1];
                }
            }
        },
        postaction=decorate
    },
    o/.default=2pt] 
\node[main] (0) {$X^0$}; 
\node[main] (1) [above right of=0]{$X^1$}; 
\node[main] (2) [below right of=1]{$X^2$}; 

    \draw[<-] (0) -- (1); 
    \draw[->] (2) -- (0); 
\end{tikzpicture}

    }
    \caption[Complete graph, unshielded triplet, collider]{An example from a fully connected triplet (a) to an unshielded triplet (b), to a collider triplet (c).}
        \label{fig:complete-unshielded}
\end{figure}
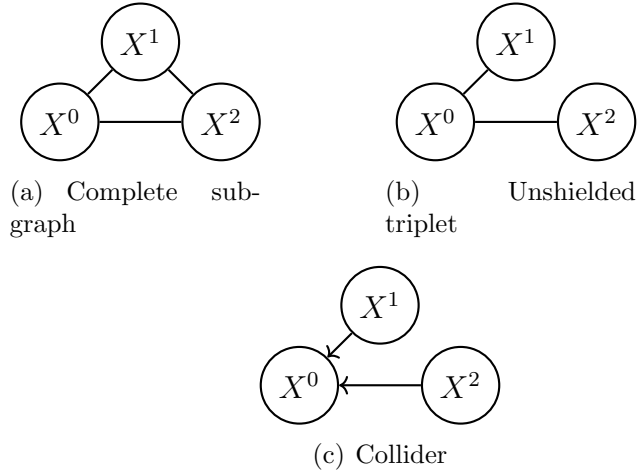

        To be more concrete, consider an initial complete graph with nodes $X^0$, $X^1$, $X^2$ of Figure \ref{fig:complete-unshielded} (a). If the edge removal phase finds that $X^1$ and $X^2$ are conditionally independent, it removes the edge between $X^1$ and $X^2$, leading to an unshielded triplet $X^0\star -\star X^1\star -\star X^2$ as sown in Figure \ref{fig:complete-unshielded} (b).  
        
In the second phase, the algorithm goes through the created unshielded triples, e.g., $X^0\star -\star X^1\star -\star X^2$, and checks if the (conditionally) independent variables ($X^0$ and $X^2$) were marked as conditionally independent while the third variable ($X^1$) was in the conditioning set. If not, the third variable ($X^1$) must be a collider, and the triple will be oriented to $X^0\star$$\rightarrow X^1 \leftarrow$$\star X^2$ as shown in Figure \ref{fig:complete-unshielded} (c).

Note that the accuracy of conditional independence tests is of paramount importance as it directly affects the edge removal phase and indirectly the edge orientation phase. 
In the next section, I will explain what affects its accuracy, as it is one of the critical strengths of the LPCMCI algorithm.

\subsection{Detection power of conditional independence tests and LPCMCI}
\label{sec:lpcmci_iterative}
The main challenge of constraint-based methods is increasing the power to detect conditional independencies, which quantifies the probability of finding a valid link between two nodes \cite{runge_pcmci_2019}.
The detection power of conditional independence-based methods can be increased by
\begin{itemize}
    \item increasing the number of samples in the dataset \cite{gerhardus_lpcmci_2020}, which results in more reliable conditional independence tests,
    \item decreasing the dimensionality of the problem \cite{runge_pcmci_2019},
    \item increasing the causal effect size between the two variables \cite{gerhardus_lpcmci_2020},
    \item increasing the statistical significance level of the conditional independence test \cite{gerhardus_lpcmci_2020}, and
    \item decreasing the size of the conditioning set until it consists only of the parents of the two variables \cite{runge_pcmci_2019}.
\end{itemize}

In real-world applications, it is often possible to take more samples, decrease the problem's dimensionality by excluding irrelevant variables through expert knowledge, and increase the effect size by enhancing the signal-to-noise ratio of the measurements.
However, these properties are less interesting when developing methods and are usually fixed through the dataset.
Furthermore, the statistical significance level alpha is commonly also determined by the false-positive rate that the researcher allows \cite{gerhardus_lpcmci_2020}.
The main focus of LPCMCI is on the last item, which means the goal is to condition on all the parents but not on other variables.

The LPCMCI algorithm improves its performance by decreasing the size of the conditioning sets by discarding conditioning sets containing known non-ancestors of the two variables of interest but including their known parents \cite{gerhardus_lpcmci_2020}. It is challenging to do so because the conditioning sets have to contain the parents of the two variables before the conditional independence tests are completed. This order is not the case for predecessor algorithms, which first conduct the conditional independence test and then orient the links. Only the orientation phase yields parenthood.
To overcome the challenge of unknown ancestries and parentships, LPCMCI entangles conditional independence tests with edge orientation to identify parentships early on. Each time the algorithm gains new knowledge about parent- or ancestorship, it updates its conditioning sets. After the algorithm converges once, it re-initializes the graph but keeps the identified parentships. 
It applies the same process but extends the conditioning sets with the previously identified parents.
This procedure should remove most false links during the second run and assign edgemarks to the remaining links. 
However, the resulting graph still tends to contain some false links between dependent variables due to latent confounding. 
To remove false links, the algorithm applies one more round of conditional independence test and edge orientations with a modified rule to select conditioning sets\footnote{For more information on these modified conditioning sets, read about the $napds_t$ sets in definition S5 of \cite{gerhardus_lpcmci_2020}.} 
to identify latently confounded links \cite{gerhardus_lpcmci_2020}. The pseudocode of this procedure can be found in Algorithm \ref{alg:orgLPCMCI}.

\begin{algorithm}
\caption{Original LPCMCI}\label{alg:orgLPCMCI}
\begin{algorithmic}[1]
\Require Observational time series dataset $\mathbf{X} = {\mathbf{X^1} , . . . , \mathbf{X}^N }$, maximal considered time lag $\tau_{max}$, significance
level $\alpha$, conditional independence test CI(X, Y, S), non-negative integer k
\State Initialize $\mathcal{C}(\mathcal{G})$ with $X_{t-\tau}^{i} \circ$$\rightarrow
X_{t}^{j}$ for $\left(0<\tau \leq \tau_{\max }\right)$ and $X_{t-\tau}^{i}  \circ$$-$$\circ X_{t}^{j} $ for $ (\tau=0)$ 
\For{$0 \leq l \leq k - 1$}
    \State \cite[Algorithm S2]{gerhardus_lpcmci_2020} Removes edges and applies orientations of $\mathcal{C}(\mathcal{G})$ via observational data $\mathbf{X}$
    \State Save all parentships marked in $\mathcal{C}(\mathcal{G})$
    \State Repeat line 1
    \Comment{Re-initilize $\mathcal{C}(\mathcal{G})$}
    \State Load parentships of line 4 and write them to $\mathcal{C}(\mathcal{G})$
\EndFor
\State Remove edges and apply orientations via observational data \cite[Algorithm S2]{gerhardus_lpcmci_2020}
\State Remove edges and apply orientations with a modified rule via observational data \cite[Algorithm S3]{gerhardus_lpcmci_2020}
\State\Return PAG $\mathcal{C}(\mathcal{G})$
\end{algorithmic}
\end{algorithm}

\subsection{Estimating effect sizes}
\label{sec:theory:effect_sizes}
So far, we learned that LPCMCI discovers causal relationships \textit{qualitatively} by returning a PAG with edges and edgemarks. 
Nevertheless, LPCMCI also returns the effect sizes of the causal relationships, which is a \textit{quantitative} measure.
LPCMCI estimates the effect sizes between two variables $X^0$ and $X^1$ with 
$\min _{\mathcal{S}} I\left(X^1; X^1 \mid \mathcal{S}\right)$, 
where $\mathcal{S}$ is a set of all conditioning sets that are being tested in the edge removal phase of node $X^0$ and $X^1$. 
This means the effect size between $X^0$ and $X^1$ is the minimum result of all conditional independence test statistic values over the conditioning sets used to test the independence between $X^0$ and $X^1$.
In the linear case, this is the lowest partial correlation between $X^0$ and $X^1$ over the tested conditioning sets $\mathcal{S}$. 
The effect sizes will become important in Chapter \ref{ch:RQTwo}, where we will use them together with the PAG to reconstruct SCMs to estimate the outcome of potential interventions.

This concludes the theoretical foundations of observational causal discovery. In the remainder of this chapter, we focus on interventional causal discovery, starting with the \textit{do}-operator.

\section{The \textit{do}-operator}
\label{sec:do}
The first research question of this thesis asks if an observational causal discovery algorithm can be improved by additionally exploiting interventional data. 
To answer this question later in this thesis, let us define the mathematical operator of interventions and the properties of interventional data.

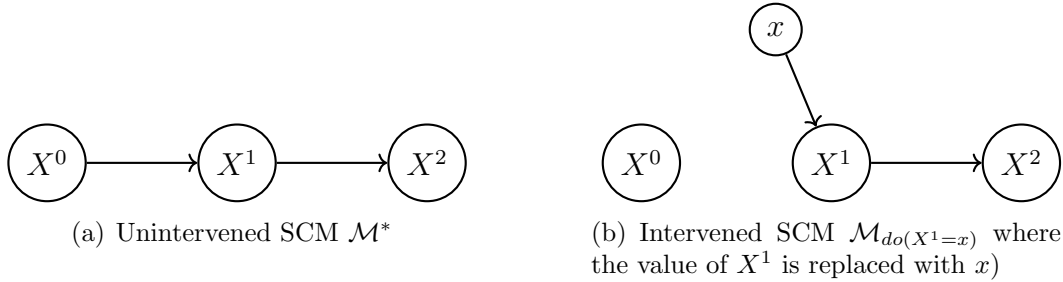
\begin{figure}[htbp]
\begin{center}
\subfigure[Unintervened SCM $\mathcal{M^*}$]{
     \begin{tikzpicture}[node distance={25mm}, thick, main/.style = {draw, circle}, hidden/.style = {draw=gray,dashed, circle}] 
        \node[main] (0) {$X^0$}; 
        \node[main] (1) [right of=0] {$X^1$}; 
        \node[main] (2) [right of=1]{$X^2$}; 
            \draw[->] (0) -- (1); 
            \draw[->] (1) -- (2); 
    \end{tikzpicture}}\hspace{15mm}
\subfigure[Intervened SCM $\mathcal{M}_{do(X^1=x)}$ where the value of $X^1$ is replaced with $x$)]{
     \begin{tikzpicture}[node distance={25mm}, thick, main/.style = {draw, circle}, hidden/.style = {draw=gray,dashed, circle}] 
        \node[main] (0) {$X^0$}; 
        \node[main] (1) [right of=0] {$X^1$}; 
        \node[main] (2) [right of=1]{$X^2$}; 
        \node[main] (3) [above right of=0]{$x$}; 
            \draw[->] (3) -- (1); 
            \draw[->] (1) -- (2); 
    \end{tikzpicture}}\hspace{5mm}

    \caption[do-operator]{Illustration of the do-operator}
    \label{fig:do}
\end{center}
\end{figure}

Consider a the original SCM $\mathcal{M^*}$ which is illustrated by \ref{fig:do}(a):
\begin{equation}
  \begin{aligned}
    &\quad X^0:= \eta_0 \\
    &\quad X^1:= f_1 (X^0 + \eta_1) \\
    &\quad X^2:= f_2 (X^1 + \eta_2).
\end{aligned}
\end{equation}
A hard intervention on a variable $X^1$ means that we set it to a chosen value $x$, irrespective of other influences that typically influence $X^1$. 
We denote this operation as $\mathit{do}(X^1=x)$, where $\mathit{do}$ is the \textit{do}-operator \cite{pearl_-calculus_2012}.
The intervened SCM $\mathcal{M}^{do(X^1=x)}$ has the structure 
\begin{equation}
  \begin{aligned}
    &\quad X^0:= \eta_0 \\
    &\quad X^1:= x \\
    &\quad X^2:= f_2 (X^1 + \eta_2), 
\end{aligned}
\end{equation}
which is also illsutrated in Figure \ref{fig:do}(b).

This type of intervention is called a \textit{hard} (or \textit{structural}) intervention in the literature \cite{eberhardt_interventions_2007}. 
Hard interventions are in contrast to \textit{soft} (or \textit{parametric}) interventions which add the intervention value to the previous value of the SCM. 
However, when I refer to just interventions, I always refer to hard interventions since I never use soft interventions in this thesis.

A distribution with a \textit{do}-operator, e.g., $P(X^2|do(X^1=x))$ is called an \textit{interventional} distribution \cite{jaber_causal_2020}.
It is important to note that this is not the same as the \textit{observational} distribution $P(X^2|X^1=x)$, which is conditioned on $X^1=x$.
The observational distribution $P(X^2|X^1=x)$ means we take only a subset of the data, where the variable $X^1$ happens to have a value of $x$.

Now that we learned about interventions and the \textit{do}-operator, we next focus on how the gold standard of interventions concerning causal discovery.

\section{Randomized control trials}
The gold standard to determine the effectiveness of interventions are studies called \textit{randomized control trials (RCTs)} that rely on a particular form of experimentation. 
In the experiments, observed and unobserved characteristics are balanced through randomization to reduce bias.
In a sufficiently large RCT, one can attribute any differences in the outcome of an experiment precisely to the intervention \cite{hariton_randomised_2018}.

Ideally, participants, including the patients, are blinded to the treatment allocations \cite{david_double-blind_2022}. 
Adequate blinding experimentally isolates the physiological effects of treatments from various psychological sources of bias.

RCT trials come at the cost of conducting experiments with many samples that must be carefully controlled by randomization, which is often tedious and expensive \cite{hariton_randomised_2018}. Furthermore, there is the possibility that the experiment's outcome has negative consequences, for example, when a proposed medication has severe side effects. Another negative aspect is that RCT trials are mainly about understanding the effect of actions but not about taking the most promising action. For example, during the COVID-19 pandemic, a group who enrolled in RCTs for vaccines did not receive the promising vaccine but a placebo.





\chapter{Causal discovery from interventional and observational data}
\label{ch:RQOne}

Recall from Section \ref{sec:lpcmci} that LPCMCI uses time ordering and m-separation to constrain the number of possible MAGs compatible with the observational data. However, even with unlimited observational data, these constraints are often insufficient to find the one true MAG representing the data-generating SCM. This is because time-ordering is not available when links are contemporaneous, and there can be multiple MAGs in the Markov equivalence class that satisfy the m-separations.

The first research question of this thesis is whether it is possible to add constraints to LCPMCI that become available through interventional data with randomized experiments to further reduce the number of MAGs that do not describe the SCM.
In the following part, I describe how to find further constraints in a dataset with interventional data points of randomized experiments.

\section{Interventional discovery}
\label{sec:intervDisc}
\begin{algorithm}
\caption{Interventional discovery}\label{alg:intervDiscov}
\begin{algorithmic}[1]
\Require 
    mixed time-series dataset $\hat{\mathbf{X}} = {\mathbf{\hat{X}^0} , \dots , \mathbf{\hat{X}}^{N-1} }$ where individual data points can be observational  ($X^i_t=f_i(pa(X^i_t))+\eta^i_t$) or interventional ($X^i_t=do(X^i_t)$),
    independence test $I(X, Y)$,
    significance level for dependency $\alpha_{\dep}$,
    significance level for independency $\alpha_{\ind}$
    
\State Initialize empty interventional dependencies list $\mathbf{L_{\dep}}$, initialize empty independencies list $\mathbf{L_{\ind}}$
\State $\mathbf{X'} \gets $subset of $ \hat{\mathbf{X}}$ where $X^i_t$ are interventional\label{ln:intervSubset} 
\State $\bar{\mathbf{X}} \gets $subset of $ \hat{\mathbf{X}}$ where $X^i_t$ are observational\label{ln:obsSubset}

\For{$0 \leq \tau \leq \tau_{max} $}
\For{$0 \leq i < N $}
\For{$0 \leq j < N $}
\If{$\tau = 0$ and $i \neq j$} \Comment{skip contemporaneous autodependencies}
    \State continue with next iteration 
\EndIf 
\State p = $I \left( \mathbf{X'}^i_{t-\tau},\bar{\mathbf{X}}^j_{t}\right)$\Comment{independency test}

\If{$p \leq \alpha_{\dep}$} \Comment{if dependent}
    \State append tuple $(j,i,\tau, p)$ to $\mathbf{L_{\dep}}$
\ElsIf{$p\geq \alpha_{\ind}$} \Comment{if independent}
    \State append tuple $(j,i,\tau, p)$ to $\mathbf{L_{\ind}}$
\EndIf
\EndFor
\EndFor
\EndFor

\While{$\mathbf{L_{\ind}}$ has a contemporaneous cycle} \Comment{ensure $\mathbf{L_{\ind}}$ is acyclic}
\State remove the item from $\mathbf{L_{\ind}}$ that has the greatest p-value and is part of the contemporaneous cycle
\EndWhile

\State\Return $\mathbf{L_{\dep}}, \mathbf{L_{\ind}}$
\end{algorithmic}
\end{algorithm}

First, I give a brief intuitive understanding of how interventional data can add constraints for causal discovery.
Then, I will elaborate in detail, and Algorithm \ref{alg:intervDiscov} gives a computational summary.
Recall from \ref{sec:rct} that experiments with a large number of randomized interventions on a variable allow attributing any differences in the observed outcome precisely to the intervention.
If interventions on variable $X$ lead to differences in the outcome of variable $Y$, we learn that $X$ can influence $Y$, and thus $X$ is an ancestor of $Y$.
On the other hand, if interventions on $X$ do not lead to differences in the outcome of variable $Y$, then $X$ is not an ancestor of $Y$.
We use this knowledge about ancestry and non-ancestry as constraints to reduce the number of MAGs compatible with the measured data but do not describe the data-generating SCM.
In the following, I describe this in more detail.

Consider a mixed time-series dataset $\hat{\mathbf{X}} = {\mathbf{\hat{X}^0}, \dots, \mathbf{\hat{X}}^{N-1} }$ where individual data points can be the result of randomized interventions ($X^i_t=do(X^i_t)$) or be observational, meaning they are the result of the unintervened process of the SCM $\mathcal{M^*}$ with $X^i_t=f_i(pa(X^i_t))+\eta^i_t$.
To determine differences in the observed outcome due to interventions, the first step is to separate the mixed time-series $\hat{\mathbf{X}}$ into a subset $\mathbf{X'}$ where all datapoints $\bar{\mathbf{X}}$ are interventional and a subset $\mathbf{\bar{X}}$ of all the observational data points.
One can then conduct independence tests between intervened variables $X'^i_{t-\tau}$ and observational variables $\bar{X}^j_{t}$ yielding p-values. 
In the case of linear relationships, we can compute the p-values of the Pearson correlation coefficient. 

If the p-value is below a significance threshold $\alpha_{\dep}$ there is reason to reject the null-hypothesis $X'^i_{t-\tau} \ind \bar{X}^j_{t}$ and accept the alternative hypothesis that $X'^i_{t-\tau} \dep \bar{X}^j_{t}$.
And since $X'^i_{t-\tau}$ is interventional but still dependent on the observational variable $\bar{X}^j_{t}$ we can conclude that $X^i_{t-\tau}$ is an ancestor of $X^j_{t}$.  I use a default significance threshold $\alpha_{\dep} = 0.05$ to reject a null hypothesis in the numerical experiments as it leads to the best results in the set of $\{0.01,0.05,0.1,0.2,0.3\}$. These results are shown in Figure \ref{fig:results_numerical}.
On the contrary, if a p-value is above the significance threshold $\alpha_{\dep}$ the null-hypothesis $X'^i_{t-\tau} \ind \bar{X}^j_{t}$ can not be rejected. 
However, failing to reject a null hypothesis does not mean that the null hypothesis $X'^i_{t-\tau} \ind \bar{X}^j_{t}$ is true \cite{johnson_insignificance_1999}.
There are methods capable of accepting and rejecting the null hypothesis. For example, Bayesian t tests computing a Bayes factor \cite{rouder_bayesian_2009}. 
Nevertheless, in this thesis, I still accept the null hypothesis based on p-values, as "Bayes factors correlate strongly with p-values"\cite{kruschke_introduction_2011}.
More precisely, I accept the null hypothesis $X'^i_{t-\tau} \ind \bar{X}^j_{t}$ if the p-value is greater than $\alpha_{\ind}$, for some $\alpha_{\ind}>\alpha_{\dep}$\footnote{Although Bayes factors correlate with p-values, it might still be better to use Bayes factors since "nonsignificant p-values obtained from null hypothesis significance testing cannot differentiate between true null effects or underpowered studies"\cite{brydges_analysis_2019}.}.
And then, since we intervened on $X^i_{t-\tau}$ but it is independent of $X^j_{t}$ we can conclude that $X^i_{t-\tau}$ is not an ancestor of $X^j_{t}$.
Note that if $\alpha_{\ind} \geq $ p-value $\geq \alpha_{\dep}$, we neither accept the null hypothesis $X'^i_{t-\tau} \ind \bar{X}^j_{t}$ nor the alternative hypothesis $X'^i_{t-\tau} \dep \bar{X}^j_{t}$. We do not gain a constraint from the interventional data in this case.
I use a default significance threshold $\alpha_{\ind} = 0.8$ to accept a null hypothesis in the numerical experiments, as it leads to best results in the set of $\{0.5,0.6,0.7,0.8,0.9,0.95\}$. These results are shown in Figure \ref{fig:results_numerical}.

\begin{figure}[htbp]
    \centering
   \begin{tikzpicture}[node distance={30mm}, thick, main/.style = {draw, circle}, hidden/.style = {draw=gray,dashed, circle}] 
\node[main] (0) {$\bar{X}^{i=0}_{t=0}$}; 
\node[main] (10) [above of=0]{$X^{\prime i=6}_{t=0}$}; 

\node[main] (1) [right of=0]{$\bar{X}^{i=0}_{t=1}$}; 
\node[main] (11) [above of=1]{$X^{\prime i=6}_{t=1}$}; 

\node[main] (2) [right of=1]{$\bar{X}^{i=0}_{t=2}$}; 
\node[main] (12) [above of=2]{$X^{\prime i=6}_{t=2}$}; 

\node[main] (3) [right of=2]{$\bar{X}^{i=0}_{t=3}$}; 
\node[main] (13) [above of=3]{$X^{\prime i=6}_{t=3}$}; 

    \draw[dotted] (0) -- (10); 
    \draw[dotted] (1) -- (11); 
    \draw[dotted] (2) -- (12); 
    \draw[dotted] (3) -- (13); 
\end{tikzpicture}

    \caption[Example independence tests]{An example of multiple time-shifted pairs of data points that are compared in the independence test $I \left( \mathbf{X'}^{i=6}_{t-(\tau=0)},\bar{\mathbf{X}}^{j=0}_{t}\right)$ }
        \label{fig:independence-timeshifted-pairs}
\end{figure}
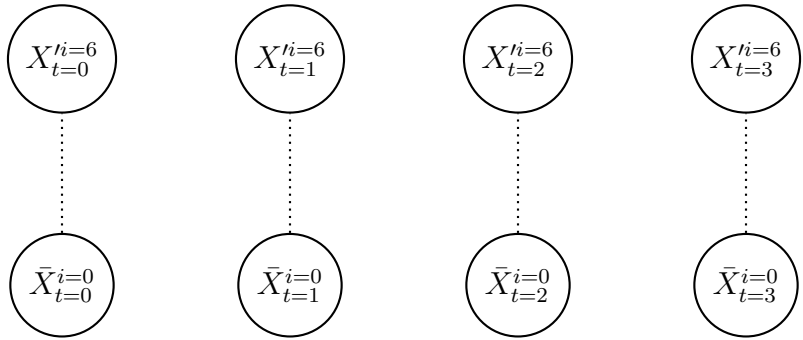

Algorithm \ref{alg:intervDiscov} conducts these independence tests for all pairs of variables $\left(\mathbf{X'}^i_{t-\tau},\bar{\mathbf{X}}^j_{t}\right)$ with $0 \leq i < N, 0 \leq j < N, 0 \leq\tau \leq \tau_{max}$, but excluding when both $i=j$ and $\tau=0$, as these are contemporaneous auto-dependencies which are incompatible with the acyclicity assumption. 
Furthermore, in this thesis, the pairs of variables $\left(\mathbf{X'}^i_{t-\tau},\bar{\mathbf{X}}^j_{t}\right)$ for the independence test is an aggregate over all time-shifted pairs $\left(X_{t-\tau}^{i}, X_{t^{}}^{j}\right)$, $\left(X_{t^{\prime}-\tau}^{i}, X_{t^{\prime}}^{j}\right)$, because their mechanisms and therefore (in)dependencies are the same due to causal stationarity.
Figure \ref{fig:independence-timeshifted-pairs} shows an example of multiple time-shifted pairs of data points that are compared in one independence test.
Algorithm \ref{alg:intervDiscov} stores the information about dependencies and independencies the lists $\mathbf{L_{\dep}}$ and $\mathbf{L_{\ind}}$, respectively. $\mathbf{L_{\dep}}$ and $\mathbf{L_{\ind}}$ can be used as additional constraints from interventions in causal discovery algorithms, as I will show next in the modified LPCMCI Algorithm \ref{alg:modLPCMCI}.

Numerical experiments (described in the next chapter) show that $97.22\%$ of the independencies found in interventional discovery are correct. However, on average, only $6.18\%$ of all the true independencies are detected through interventional discovery.
This ratio of $6.18\%$ to $97.22\%$ is surprisingly extreme and influenced by $\alpha_{\ind}$. As mentioned earlier, $\alpha_{\ind}$ is set to $0.8$ as it leads to the best results. This indicates that falsely detected independencies tend to have a strong negative impact. This is indeed the case, as false independence constraints from interventional discovery can not be corrected through LPCMCI, as explained in the next section.

Furthermore, $66.30\%$ of the dependencies found in interventional discovery are correct. And on average, $48\%$ of the true dependencies are detected through interventional discovery.

\section{Observational causal discovery with interventional constraints}
\label{sec:modLPCMCI}
We learned how to get causal constraints from interventional data in the last section.
This section explains how the LPCMCI algorithm (Algorithm \ref{alg:orgLPCMCI}), which is originally a purely observational causal discovery algorithm, is extended to profit from such additional causal constraints. 
Separating the process of getting causal constraints and using them to reconstruct the causal graph has the advantage of modularity.
Indeed, where these additional causal constraints come from does not matter. 
The only requirement is that the causal constraints are information about ancestry or non-ancestry. 
However, in this thesis, the additional causal constraints come from the interventional causal discovery algorithm (Algorithm \ref{alg:intervDiscov}). 
In the following, I will first give a short intuitive overview of the modified LPCMCI algorithm and will then describe it in more detail. Furthermore, the pseudocode is in Algorithm \ref{alg:modLPCMCI}.

As described in Section \ref{sec:lpcmci}, the original LPCMCI algorithm operates in a loop. Inside the loop, it initializes its working memory graph $\mathcal{C}(\mathcal{G})$ with a complete graph except for time constraints; if available, it writes the known ancestors from the last iteration to $\mathcal{C}(\mathcal{G})$ and starts its observational causal discovery process by removing and orienting edges in $\mathcal{C}(\mathcal{G})$ and additionally saves found ancestors outside of $\mathcal{C}(\mathcal{G})$.
The main modification to LPCMCI is that we now apply the constraints from interventional discovery right after the initialization of $\mathcal{C}(\mathcal{G})$ and just before the ancestors from the last iteration are written. 

The original LPCMCI algorithm assumes that the time-series dataset is purely observational, where every datapoint $X^i_t$ is the result of the unintervened process of the SCM with $X^i_t=f_i(pa(X^i_t))+\eta^i_t$.
However, now the time-series $\hat{\mathbf{X}} = {\mathbf{\hat{X}^1}, \dots, \mathbf{\hat{X}}^N }$ can be mixed, meaning that there are still some observational data points $\left(X^i_t=f_i(pa(X^i_t))+\eta^i_t\right)$, but there may also be interventional datapoints $\left(X^i_t=do(X^i_t)\right)$.
Similarly to Algorithm \ref{alg:intervDiscov} Lines \ref{ln:intervSubset}-\ref{ln:obsSubset} we take the subset $\mathbf{X'}$ of the mixed time-series $\hat{\mathbf{X}}$ that comprises all interventional datapoints $\mathbf{X'}$
and a subset $\mathbf{\bar{X}}$ that comprises all the observational data points 
$\mathbf{\bar{X}}$.
After the working memory $\mathcal{C}(\mathcal{G})$ is initialized, the information of the interventional independencies list $\mathbf{L_{\dep}}$ is used.
Each element in $\mathbf{L_{\dep}}$ contains a tuple $(j,i,\tau, p)$ encoding that 
$X'^i_{t-\tau} \ind \bar{X}^j_{t}$ which means $X^i_{t-\tau} \notin an(X^j_{t})$. Since the arrowhead edgemarks ($<,>$) encode non-ancestry, we write 
$X^i_{t-\tau} \leftarrow$$\star X^j_{t}$. Note that $X'^i_{t-\tau} \ind \bar{X}^j_{t}$ does not give any information about whether $\bar{X}^j_{t}$ is an ancestor of $X^i_{t}$ or not. Hence, its edgemark will not be changed. An unchanged edgemark edgemark, whatever it is, is denoted by a star edgemark ($\star$).

Next we include the constraints of the independencies list $\mathbf{L_{\dep}}$ each element contains tuple $(j,i,\tau, p)$ encoding that $X'^i_{t-\tau} \dep \bar{X}^j_{t}$.
Since $X'^i_{t-\tau}$ is interventional but still dependent on the observational variable $\bar{X}^j_{t}$, we can conclude that $X^i_{t-\tau} \in an\left(X^j_{t}\right)$
(and $X^j_{t} \in de\left(X^i_{t-\tau}\right)$). Acyclicity tells us that $X^j_{t} \notin an\left(X^i_{t-\tau}\right)$.
Therefore, we constrain $\mathcal{C}(\mathcal{G})$ with $X^i_{t-\tau} \rightarrow X^j_{t}$.
Note that the known interventional independencies never remove an edge between a pair of variables $X^i_{t-\tau}$ and $X^j_{t}$, even if $X'^i_{t-\tau} \ind \bar{X}^j_{t}$ and $X'^j_{t-\tau} \ind \bar{X}^i_{t}$. In this case $X^i_{t} \notin an\left(X^j_{t-\tau}\right)$ and $X^j_{t} \notin an\left(X^i_{t-\tau}\right)$, leading to $X^j_{t} \leftrightarrow X^i_{t-\tau}$, meaning $X^j_{t}$ and $X^i_{t-\tau}$ can still be associated due to at least one (latent) confounder $\left(X^j_{t} \leftarrow \left\{V^k_t, ...\right\} \rightarrow X^i_{t-\tau}\right)$.
If $X^i_{t-\tau}$ and $X^j_{t}$ are not associated, we rely on a conditional independence test of the next step to remove that edge.

After applying the external constraints in $\mathbf{L_{\dep}}$ and $\mathbf{L_{\ind}}$ (in our case from interventional causal discovery) to $\mathcal{C}(\mathcal{G})$, the edges are removed and oriented just like in the original LPCMCI algorithm, using the observational data $\mathbf{\bar{X}}$. 
In the following, I describe what this means practically. When external constraints indicate that $X^0 \in an(X^1)$, the link is written as $X^0 \rightarrow X^1$ in the working memory. 
             
Such edge removals can be necessary to remove a link when $X^0 \in an(X^1)$ but $X^0$ is not a direct parent of $X^1$ ($X^0 \notin pa(X^1)$).
In numerical experiments, I observe that, on average, $45\%$ of the links written due to interventional discovery survive LPCMCI. 

Unfortunately, it is also possible that an edge written due to an interventional constraint is again removed through a \textit{falsely} detected conditional independence in the observational data. This is a limitation of my proposed method. In numerical experiments, I observe that, on average, $55\%$ of the \textit{correct} links written due to interventional discovery survive LPCMCI. Meaning $45\%$ of the correctly discovered links are lost again. 
It could be useful to try to limit LPCMCI's ability to remove external dependency constraints in future work.
Similarly, when external constraints indicate that $X^0 \notin an(X^1)$, the link is written as $X^0 \leftarrow$$\circ X^1$ in the working memory. LPCMCI can still modify entry by removing the link ($X^0 \hspace{3mm} X^1$), write an association due to confounding ($X^0 \leftrightarrow X^1$), or write that $X^1$ is an ancestor of $X^0$ ($X^0 \leftarrow X^1$). In empirical experiments, I observe $100\%$ of the independencies detected through interventional discovery survive LPCMCI, which is expected as LPCMCI does not have the ability to add links. 

\begin{figure}[htbp]
    \centering
        \subfigure[Original output of LPCMCI without additional constraints from interventional causal discovery. Note that there is a $4 \leftarrow$$\circ 0$ link. This link \textit{allows} the possibility that $X^0 \in an(X^4)$.]{
    \includegraphics[width=0.85\linewidth]{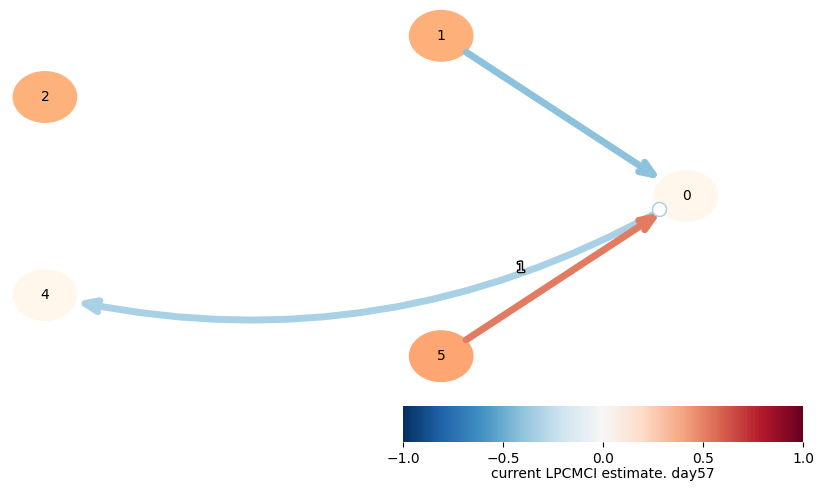}}\hspace{5mm}
             
    \subfigure[Output of the modified LPCMCI with the external information that node $X^0 \ind X^4$ when intervening on $X^0$. The modified LPCMCI algorithm now writes a $4 \leftrightarrow 0$ link. This link \textit{excludes} the possibility that node $X^0 \in an(X^4)$.]{
    \includegraphics[width=0.85\linewidth]{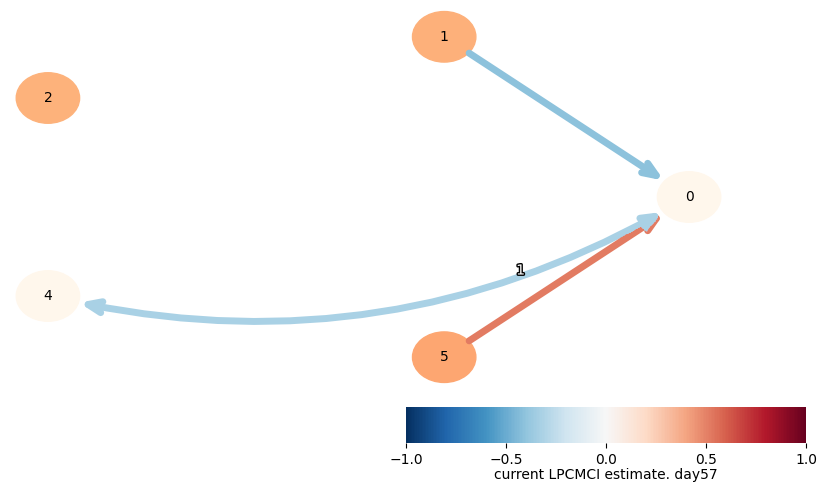}}
    \caption[Using external constraints]{An example of how the modified LPCMCI algorithm successfully uses the external information that node $0$ is not an ancestor of node $4$.}
        \label{fig:ext-indep}
\end{figure}

Interventional constraints can improve the resulting graph of LPCMCI directly when link orientations can not be discovered through observational data.
Figure \ref{fig:ext-indep} shows an example of how the modified LPCMCI algorithm successfully uses external information about a non-ancestorship that could not be detected through purely observational causal discovery.

In addition to the direct benefits of interventional constraints, there are indirect benefits to observational causal discovery because there is the opportunity for more accurate conditional independence tests.
As explained in Section \ref{sec:lpcmci}, conditional independence tests $CI(X, Y,\mathbf{S})$ are more accurate when the conditioning sets $\textbf{S}$ consist only of the closest ancestors of $X, Y$, ideally their parents. 
Additional information about (non-)ancestries from interventional data allows LPCMCI to remove known non-ancestors of $X$ and $Y$ from $\textbf{S}$ and add the known parents of $X$ or $Y$ to $\textbf{S}$ more accurately.

As mentioned earlier, this extended LPCMCI algorithm uses only the observational data $\mathbf{\bar{X}}$ of conditional independence tests.
However, it may be possible to include some interventional data to increase its robustness. I elaborate on this idea in Section \ref{sec:interventionalDataInLPCMCI}, but its implementation is beyond the scope of this thesis.

This concludes the description of how I detect causal constraints in interventional data and how I extended the LPCMCI algorithm to exploit such external causal constraints.
The next chapter explains how interventional and observational causal discovery can be used for regret-minimizing control.

\begin{algorithm}
\caption{Modified LPCMCI}\label{alg:modLPCMCI}
\begin{algorithmic}[1]
\Require 
    mixed time-series dataset $\hat{\mathbf{X}} = {\mathbf{\hat{X}^1} , \dots , \mathbf{\hat{X}}^N }$ where individual data points can be observational  ($X^i_t=f_i(pa(X^i_t))+\eta^i_t$) or interventional ($X^i_t=do(X^i_t)$),
    interventional dependencies list $\mathbf{L_{\dep}}=[(j,i,\tau,\_),...]$ indicating $X^i_{t-\tau}\in an(X^j_{t})$ and $X^j_{t} \in de(X^i_{t-\tau})$,
    interventional independencies $\mathbf{L_{\ind}}=[(j,i,\tau,\_),...]$ indicating $X^i_{t-\tau} \notin an(X^j_{t})$,
    maximal considered time lag $\tau_{max}$, 
    significance level $\alpha$, 
    CI test CI(X, Y, S), 
    non-negative integer k 
\State $\bar{\mathbf{X}} \gets $subset of $ \hat{\mathbf{X}}$ where $X^i_t$ are observational

\State Initialize $\mathcal{C}(\mathcal{G})$ with $X_{t-\tau}^{i} \circ$$\rightarrow
X_{t}^{j}$ for $\left(0<\tau \leq \tau_{\max }\right)$ and $X_{t-\tau}^{i}  \circ$$-$$\circ X_{t}^{j} $ for $ (\tau=0)$

\For{ $(j,i,\tau, p)$ dependency in $\mathbf{L_{\dep}}$} \label{ln:write_interv_start}
    \State  write edgemark $X^i_{t-\tau} \rightarrow X^j_{t}$ 
\EndFor

\For{ $(j,i,\tau, p)$ independency in $\mathbf{L_{\ind}}$}
    \State  write edgemark $X^i_{t-\tau} \leftarrow$$\star$$ X^j_{t}$ 
\EndFor \label{ln:write_interv_stop}

\For{$0 \leq l \leq k - 1$}
    \State Remove edges and apply orientations of $\mathcal{C}(\mathcal{G})$ via observational data \cite[Algorithm S2]{gerhardus_lpcmci_2020}
    \State Save all ancestorships $X^i_{t-\tau} \in an(X^j_{t})$ marked in $\mathcal{C}(\mathcal{G})$ \label{ln:anc}
    \State Repeat line 2\Comment{Re-initilize $\mathcal{C}(\mathcal{G})$}
    \State Load ancestorships of line \ref{ln:anc} and write them to $\mathcal{C}(\mathcal{G})$
    \State Repeat line \ref{ln:write_interv_start}-\ref{ln:write_interv_stop} \Comment{write interventional (in)dependencies in $\mathbf{L_{\dep}} $ and $\mathbf{L_{\ind}}$ to $\mathcal{C}(\mathcal{G})$}
\EndFor
\State Remove edges and apply orientations via observational data \cite[Algorithm S2]{gerhardus_lpcmci_2020}
\State Remove edges and apply orientations with a modified rule via observational data\cite[Algorithm S3]{gerhardus_lpcmci_2020}
\State\Return PAG $\mathcal{C}(\mathcal{G})$ with its effect sizes
\end{algorithmic}
\end{algorithm}

\chapter{Causal learning for regret-minimizing optimal control}
\label{ch:RQTwo}
\section{Overview}

The second research question of this thesis is whether LPCMCI with interventional constraints can be used to learn a causal model, take the best action with respect to the model and improve the model with the resulting interventional data of the action. 
I propose an algorithm to answer this research question and evaluate its performance in a simulation study.
First, I will give the macrostructure of the simulation study and will then explain the missing details.
The pseudocode of the macrostructure can be found in Algorithm \ref{alg:loop}.

Since it is a simulation study, we simulate the data-generating environment we want to understand and control. 
Therefore I start by sampling a data-generating SCM that generates an initial observational time-series $ \mathbf{V}=\left\{\mathbf{V}^{0}, \ldots, \mathbf{V}^{\tilde{N}-1}\right\}$ with $\mathbf{V^i}=\left\{{V}^i_{0},  \ldots, {V}^i_{T_{init}-1}\right\}$ and a length of $T_{init}$ samples.
We generate a length of $T_{init}$ observational samples because, as described in the introduction, it is common that one first observes the environment for a while before trying to control it. In my numerical experiments, $T_{init}=50$ if not otherwise mentioned.
The details of this data generation will be explained in the next section.
I then simulate the presence of latent variables by dropping $N-\tilde{N}$ variables from $ \mathbf{V}=\left\{\mathbf{V}^{0}, \ldots, \mathbf{V}^{\tilde{N}-1}\right\}$ which yields the measured time-series $\mathbf{X}=\left\{\mathbf{X}^{0}, \ldots, \mathbf{X}^{N-1}\right\}$ with $N \leq \tilde{N}$.

If there are interventional data available, Algorithm \ref{alg:intervDiscov} searches for (in)dependencies in the interventional data as described earlier in Section \ref{sec:intervDisc}.
Then the modified LPCMCI Algorithm (previously introduced as Algorithm \ref{alg:modLPCMCI}) estimates the causal PAG $\mathcal{G}$ from the interventional (in)dependencies and observational data. 
The causal PAG $\mathcal{G}$, together with its effect sizes, are used to reconstruct possible SCMs over the observed variables that created the data.
The control Algorithm \ref{alg:do_opt}, described later in Section \ref{sec:control}, uses these SCMs and proposes an intervention $do(X^{i_{opt}}_t=x)$ that aims to optimize the target variable $X^y$ and/or explores causal relationships by gathering interventional data. In the numerical experiments, the intervention is only executed every fourth time. The other three out of four times are purely observational steps.

We then intervene in the SCM $\mathcal{M}$ with the proposed intervention $do(i_{opt}=x)$ and simulate the environment by generating the state $\mathbf{V}_{T_{init}+t}$ of the next timestep, again with Algorithm \ref{alg:data-generato}.
After the next time step is generated, we evaluate the control performance by computing the induced regret, which is the difference between the outcome of the actual intervention and the theoretically optimal intervention. Section \ref{sec:average_regret} describes this evaluation process in more detail.

The algorithm then iterates and runs this loop as long as $t<T_{max}+T_{ini}$. In the numerical experiments, $T_{max}=200$ if not explicitly changed.


\begin{algorithm}
\caption{Simulating optimal control with causal discovery}\label{alg:loop}
\begin{algorithmic}[1]
\Require SCM $\mathcal{M}$, index of the target variable to optimize $y$
\State Algorithm \ref{alg:data-generato} simulates the environment by generating the first $T_{init}$ samples of the time-series $ \mathbf{V}=\left\{\mathbf{V}^{0}, \ldots, \mathbf{V}^{\tilde{N}-1}\right\}$ with $\mathbf{V}^i=\left\{{V}^i_{0},  \ldots, {V}^i_{T_{init}-1}\right\}$
\For{$0 \leq t < T_{max}$} 
    \State Drop the latent variables of $\mathbf{V}=\left\{\mathbf{V}^{0}, \ldots, \mathbf{V}^{\tilde{N}-1}, \right\}$ which yields the measured time series $\mathbf{X}=\left\{\mathbf{X}^{0}, \ldots, \mathbf{X}^{N-1}\right\}$ with $N \leq \tilde{N}$ \label{ln:measurements}
    \State Get interventional dependencies $\mathbf{L_{\not\!\perp\!\!\!\perp}}$ and interventional independencies $\mathbf{L_{\!\perp\!\!\!\perp}}$ via Algorithm \ref{alg:intervDiscov}
    \State Discover the causal PAG $\mathcal{G}$ with its effect sizes given $\mathbf{X}, \mathbf{L_{\not\!\perp\!\!\!\perp}}, \mathbf{L_{\!\perp\!\!\!\perp}}$ with Algorithm \ref{alg:modLPCMCI}
    \State Find the optimal intervention $do(X^{i_{opt}}=x_{opt})$ to optimize target variable $X^y$ given $\mathcal{G}$ with its effect sizes, using Algorithm \ref{alg:do_opt}
    \State $x \gets  \sim{Bin}\left(1, \epsilon, \left\{x_{opt}, x_{alt}\right\}\right)$ \Comment{for more details see explanation of Equation \ref{eq:exploration-exploitation}}
    \State Intervene in $\mathcal{M}$ with $do(i_{opt}=x$)  and simulate the environment by generating the next time-step $\mathbf{V}_{T_{init}+t}$ with Algorithm \ref{alg:data-generato} 
    \State Evaluate the performance by computing the induced regret with Equation \ref{eq:average_regret}

\EndFor
\end{algorithmic}
\end{algorithm}

In the following sections, I describe the mentioned methods, except for the already known Algorithms \ref{alg:modLPCMCI} and \ref{alg:intervDiscov}.
I start by explaining the sampling of SCMs, which generate the synthetic data.

\section{Sampling SCMs that simulate the environment}
\label{sec:sampling_scm}
The original motivation of this thesis is to discover the causal dependencies between a person's mood and other variables that are easy to measure via consumer services and devices like fitness trackers for sleep and exercise or weather apps to recommend interventions that lead to better outcomes \cite{reiser_predicting_2022} \cite{reiser_causal_2022}. 
However, it is difficult to evaluate the causal graph of the real-world datasets as the ground truth is unknown. Therefore, I sample synthetic SCMs that generate time series with similar properties compared to the real-world dataset.

\begin{figure}[htbp]
\centering
\includegraphics[width=0.8\linewidth]{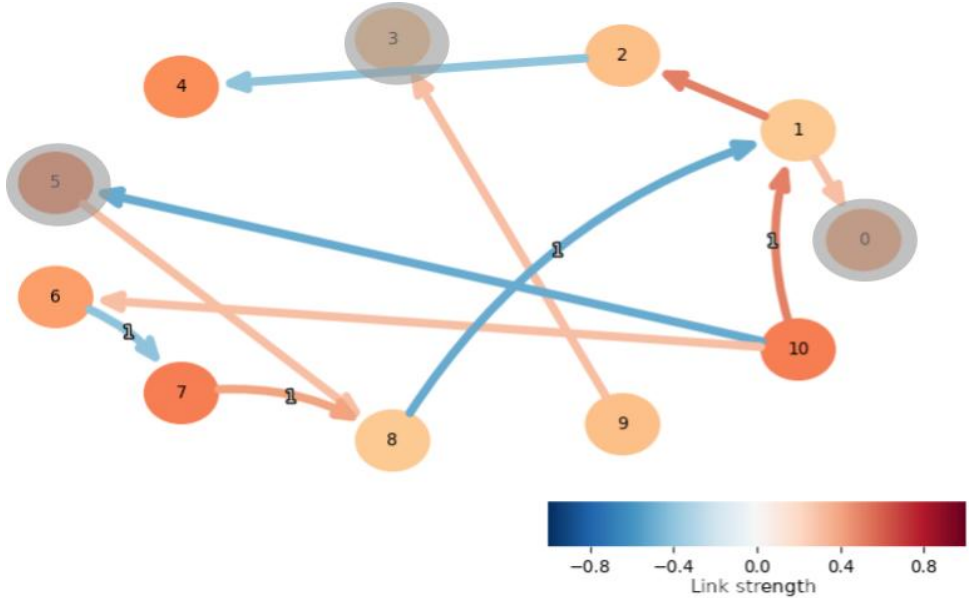}

    \caption[Example SCM]{The DAG of a sample SCM with $\tilde{N}=11$ auto-dependent variables with eleven cross-dependencies. The number in the nodes represents the index $i$ of a variable $X^i$. All auto-dependencies have a time delay of $\tau=1$. A "1" written on a curved arrow indicates that the cross-dependency has a time delay $\tau=1$. Straight arrows without a number represent contemporaneous effects ($\tau=0$). The color of an edge or node represents the strength of the cross- or auto-dependency, respectively.}
        \label{fig:3_latent}
\end{figure}

A data-generating SCMs are similar to the ones used for my research module \cite{reiser_causal_2022}, and their equations are given by
\begin{equation}
\label{eq:scm_data_gen}
V_{t}^{j}:=a_{j} V_{t-1}^{j}+\sum_{i} c_{i} f_{i}\left(V_{t-\tau_{i}}^{i}\right)+\eta_{t}^{j} \quad \text { for } \quad t \in\{0, \ldots, T-1\}, \quad j \in\{0, \ldots, \tilde{N-1} \}.
\end{equation}
The model consists of $\tilde{N}$ variables $V_t^j$ of which $N$ variables $X_t^j$ are selected randomly to be measured. The other $\tilde{N}-N$ variables are latent.
In the numerical experiments, the default parameter settings are $\tilde{N}=7$ and $N=5$.
Every variable $V_t^j$ has a linear auto-dependency $a_j V_{t-1}^{j}$, with $a_j \sim \mathcal{U}(0.3,0.6)$, meaning that the strength of the auto-dependency is drawn from a uniform distribution with a lower bound of 0.3 and an upper bound of 0.6.
In total there are $L=\tilde{N}$ randomly chosen variable pairs $(V_t^j,V^i_{t-\tau_i})$ with non-zero linear cross-dependencies $f_i \sim \pm \mathcal{U}(0.2,0.5)$. Otherwise, $f_i=0$, meaning that all other variables have no cross-dependency.
60\% of the links are instantaneous, meaning they have a time lag of $\tau =0$. The other 40\% have a time lag of $\tau=1$.
The noises $\eta^{j} \sim \mathcal{N}(0,\mathcal{U}[0.5,2])$.
I repeat each parameter set in the experiments for 500 different SCM samples. If an SCM generates a non-stationary time series, it is redrawn. This can happen when feedback loops in the SCM lead to trends or periodic patterns in the values or variance.
Furthermore, an SCMs is also redrawn when there is no cross-dependency influencing the target variable $X^y$, as $X^y$ would then be uncontrollable except for the trivial case of intervening directly on the target variable $X^y$.
Figure \ref{fig:3_latent} shows an example of a sampled SCM.

The following section explains how SCMs can be used to generate data.

\section{Generating synthetic data with SCMs}
\label{sec:data-gen}
In this thesis, generating data with SCMs has two use cases.
In the first use-case, the SCM is sampled as just described in Section \ref{sec:sampling_scm} and used to generate synthetic data that mimics data from the environment. This is only needed because the numerical experiments rely on synthetic data. This synthetic data is not needed in a real-world application because real measurements are used.
The second use-case of generating data is to simulate the outcome of hypothetical interventions to find the optimal intervention. 
Note that the use-case of generating synthetic data that mimics data from the environment includes all variables $V^i$ with $i=0,\dots,\tilde{N}-1$, including the latents.
In contrast, in the second use-case of simulating the outcome of hypothetical interventions to find the optimal intervention, only the observed variables $X^i$ with $i=0,\dots,{N}-1$ are generated as the latents are unknown.
The following algorithm can generate data for both use cases. However, in the notation, I use $V$ to describe the variables and $\tilde{N}$ for their number. Please note that this should actually be an $X$ and ${N}$ if the latents are not included.

\begin{algorithm}
\caption[Synthetic data generation]{Generate an (interventional) time-series given an SCM (and a starting sequence)}\label{alg:data-generato}
\begin{algorithmic}[1]
\Require 
SCM $\mathcal{M}$, 
intervention $do(V^i=x)$, 
Starting sequence $\mathbf{V}_{start}$ of length $\tau_{max}$,
number of samples to generate $T_{gen}$ \label{ln:req}
\State $V^{i_{interv}} \gets$ the intervention variable from $do(V^i=x)$
\State $\tilde{N} \gets$ number of variables in $\mathcal{M}$
\State Initialize the starting sequence $\mathbf{V}_{t=0 ,\dots ,
\tau_{max}-1} \gets \mathbf{V_{start}}$
\For{$\tau_{max} \leq t < T_{gen} + \tau_{max}$}
    \For{all variables $V^i$ in $\tilde{N}$ sorted topologically by the causal order of the contemporaneous graph of $\mathcal{M}$}
        \If{$V^i = V^{i_{interv}}$} \Comment{Intervention}
            \State $V_{t}^{i} \gets do(V^i=x)$ 
        \ElsIf{$V^i \neq V^{i_{interv}}$} \Comment{No intervention}
            \State $V_{t}^{i} \gets \sim \mathcal{N}(0,\mathcal{U}[0.5,2])$ \Comment{Noise}
            \For{$0 \leq j < \tilde{N}$}
                \State $f^{i,j}, \tau  \gets$ mechanism $f^{i,j}$ of $\mathcal{M}$ with the information how variable $X^j$ affects $X^i$ with time lag $\tau$
                \State $V_{t}^{i} \gets V_{t}^{i} + f^{i,j}(V_{t - \tau}^j)$ \Comment{Add effect of $j$ on $i$ with time lag $\tau$}
            \EndFor
        \EndIf
    \EndFor
\EndFor
\State $\mathbf{V} \gets \mathbf{V}_{t=\tau_{max}, \dots ,\tau_{max}+ T_{gen}}$ \Comment{remove starting sequence of length $\tau_{max}$}
\State\Return generated time-series $\mathbf{V}$
\end{algorithmic}
\end{algorithm}

\subsection{Initialization and warm up}
When we generate a synthetic time series and the maximal time lag $\tau_{max}>0$, we need a starting sequence of length $\tau_{max}$ to generate new data.
In our numerical experiments, this starting sequence is determined by the random noises $\eta^{j}_t$. A problem that arises due to the random initialization is that the first generated values heavily depend on the initial values, which come from a different distribution, inducing an initialization bias. 
After $50$ time steps, I observed that the values converged to their normal range. Therefore, to remove this initialization bias, I use the $50^{th}$ data sample as the actual initialization value of the SCM. 
This warm-up phase is only used when the initialization values come from a different distribution.

In contrast, during the optimal control simulation, it is necessary to continue generating the next timestep from previous samples.
If there is a time lag $\tau>0$ in the SCM, the data generator requires a starting sequence $\mathbf{V}_{start}$ which are the last $\tau_{max}$ samples of the time-series to start from.
There is no warm-up period when we want to continue generating a time series from a given starting sequence.

\subsection{Observational and interventional data generation}
We need to be able to simulate both observational and interventional data.
Observational data-generation is determined by the SCM of Equation \ref{eq:scm_data_gen} generating a timeseries $ \mathbf{V}=\left\{\mathbf{V}^{0}, \ldots, \mathbf{V}^{\tilde{N}-1}\right\}$ with $\mathbf{V^i}=\left\{{V}^i_{0},  \ldots, {V}^i_{T_{max}-1}\right\}$ with $\tilde{N}$ variables and a length of $T_{max}$. 
To generate interventional data, the original SCM (Equation \ref{eq:scm_data_gen}) is modified through an intervention $do(V^i = x)$.

This concludes the explanation of data generation with SCMs. The next section describes the control algorithm that recommends an intervention.


\section{Control}
\label{sec:control}

The control algorithm emulates conducting interventions of randomized control trials (RCTs).
Recall from Section \ref{sec:rct} that an RCT compares a new treatment to an alternative treatment.
The goal of this control algorithm is to use a PAG $\mathcal{G}$ to assign treatments, i.e., interventions according to 
\begin{equation}
\label{eq:do}
    do\left(X^{i_{opt}}=x\right), 
    \end{equation}
    with
\begin{equation}
\label{eq:exploration-exploitation}
    x\sim{Bin}\left(1, \epsilon, \left\{x_{opt}, x_{alt}\right\}\right), 
    \end{equation}

where ${Bin}\left(1, 1-\epsilon, \left\{x_{opt}, x_{alt}\right\}\right)$ is the outcome of one binomial experiment with the probability of $1-\epsilon$ of taking the optimistic intervention value and a probability of $\epsilon$ to take the alternative intervention value.

This means the intervention has $1-\epsilon$ chance to be the most optimistic intervention $do(X^{i_{opt}}=x_{opt})$, where we intervene on the optimal variable $X^{i_{opt}}$ and its value is set to the most optimistic value $x_{opt}$. 
With a probability of $\epsilon$, we still intervene with $do(X^{i_{opt}}=x_{alt})$ where we still intervene on the optimal variable $X^{i_{opt}}$ but with an alternative intervention value $x_{alt}$. 

\begin{figure}[htbp]
\begin{tikzpicture}
\begin{axis}[
    title={$\epsilon(t)$},
    xlabel={Timestamp},
    ylabel={$\epsilon$},
    xmin=0, xmax=200,
    ymin=0, ymax=0.5,
    xtick={0,50,100,150,200},
    ytick={0,0.1,0.2,0.3000,0.4000,0.5000},
    legend pos=south west,
    ymajorgrids=true,
    grid style=dashed,
]
\addplot[
    color=blue,
    mark=none,
    ]
    coordinates {
 (0,0.5)(1,0.495)(2,0.49005)(3,0.4851495)(4,0.480298005)(5,0.47549502495)(6,0.4707400747005)(7,0.466032673953495)(8,0.46137234721396)(9,0.45675862374182)(10,0.452191037504402)(11,0.447669127129358)(12,0.443192435858065)(13,0.438760511499484)(14,0.434372906384489)(15,0.430029177320644)(16,0.425728885547438)(17,0.421471596691963)(18,0.417256880725044)(19,0.413084311917793)(20,0.408953468798615)(21,0.404863934110629)(22,0.400815294769523)(23,0.396807141821828)(24,0.392839070403609)(25,0.388910679699573)(26,0.385021572902578)(27,0.381171357173552)(28,0.377359643601816)(29,0.373586047165798)(30,0.36985018669414)(31,0.366151684827199)(32,0.362490167978927)(33,0.358865266299137)(34,0.355276613636146)(35,0.351723847499785)(36,0.348206609024787)(37,0.344724542934539)(38,0.341277297505194)(39,0.337864524530142)(40,0.33448587928484)(41,0.331141020491992)(42,0.327829610287072)(43,0.324551314184201)(44,0.321305801042359)(45,0.318092743031935)(46,0.314911815601616)(47,0.3117626974456)(48,0.308645070471144)(49,0.305558619766433)(50,0.302503033568768)(51,0.299478003233081)(52,0.29648322320075)(53,0.293518390968742)(54,0.290583207059055)(55,0.287677374988464)(56,0.28480060123858)(57,0.281952595226194)(58,0.279133069273932)(59,0.276341738581193)(60,0.273578321195381)(61,0.270842537983427)(62,0.268134112603593)(63,0.265452771477557)(64,0.262798243762781)(65,0.260170261325153)(66,0.257568558711902)(67,0.254992873124783)(68,0.252442944393535)(69,0.249918514949599)(70,0.247419329800103)(71,0.244945136502102)(72,0.242495685137081)(73,0.240070728285711)(74,0.237670021002854)(75,0.235293320792825)(76,0.232940387584897)(77,0.230610983709048)(78,0.228304873871957)(79,0.226021825133238)(80,0.223761606881905)(81,0.221523990813086)(82,0.219308750904955)(83,0.217115663395906)(84,0.214944506761947)(85,0.212795061694327)(86,0.210667111077384)(87,0.20856043996661)(88,0.206474835566944)(89,0.204410087211275)(90,0.202365986339162)(91,0.20034232647577)(92,0.198338903211013)(93,0.196355514178903)(94,0.194391959037113)(95,0.192448039446742)(96,0.190523559052275)(97,0.188618323461752)(98,0.186732140227135)(99,0.184864818824863)(100,0.183016170636615)(101,0.181186008930248)(102,0.179374148840946)(103,0.177580407352537)(104,0.175804603279011)(105,0.174046557246221)(106,0.172306091673759)(107,0.170583030757021)(108,0.168877200449451)(109,0.167188428444957)(110,0.165516544160507)(111,0.163861378718902)(112,0.162222764931713)(113,0.160600537282396)(114,0.158994531909572)(115,0.157404586590476)(116,0.155830540724571)(117,0.154272235317326)(118,0.152729512964152)(119,0.151202217834511)(120,0.149690195656166)(121,0.148193293699604)(122,0.146711360762608)(123,0.145244247154982)(124,0.143791804683432)(125,0.142353886636598)(126,0.140930347770232)(127,0.139521044292529)(128,0.138125833849604)(129,0.136744575511108)(130,0.135377129755997)(131,0.134023358458437)(132,0.132683124873853)(133,0.131356293625114)(134,0.130042730688863)(135,0.128742303381974)(136,0.127454880348155)(137,0.126180331544673)(138,0.124918528229226)(139,0.123669342946934)(140,0.122432649517465)(141,0.12120832302229)(142,0.119996239792067)(143,0.118796277394147)(144,0.117608314620205)(145,0.116432231474003)(146,0.115267909159263)(147,0.11411523006767)(148,0.112974077766994)(149,0.111844336989324)(150,0.11072589361943)(151,0.109618634683236)(152,0.108522448336404)(153,0.10743722385304)(154,0.106362851614509)(155,0.105299223098364)(156,0.104246230867381)(157,0.103203768558707)(158,0.10217173087312)(159,0.101150013564389)(160,0.100138513428745)(161,0.0991371282944572)(162,0.0981457570115127)(163,0.0971642994413975)(164,0.0961926564469836)(165,0.0952307298825137)(166,0.0942784225836886)(167,0.0933356383578517)(168,0.0924022819742732)(169,0.0914782591545304)(170,0.0905634765629851)(171,0.0896578417973553)(172,0.0887612633793817)(173,0.0878736507455879)(174,0.086994914238132)(175,0.0861249650957507)(176,0.0852637154447932)(177,0.0844110782903453)(178,0.0835669675074418)(179,0.0827312978323674)(180,0.0819039848540437)(181,0.0810849450055033)(182,0.0802740955554483)(183,0.0794713545998938)(184,0.0786766410538948)(185,0.0778898746433559)(186,0.0771109758969223)(187,0.0763398661379531)(188,0.0755764674765736)(189,0.0748207028018078)(190,0.0740724957737898)(191,0.0733317708160519)(192,0.0725984531078913)(193,0.0718724685768124)(194,0.0711537438910443)(195,0.0704422064521339)(196,0.0697377843876125)(197,0.0690404065437364)(198,0.068350002478299)(199,0.067666502453516)(200,0.0669898374289809)
    };
\end{axis}
\end{tikzpicture}
    \caption[Exploration decay]{Exploration decay: The alternative intervention is initially taken with a probability of 0.5. This probability decays exponentially and reaches 0.067 after 200 interventions.}
    \label{fig:eps}
\end{figure}
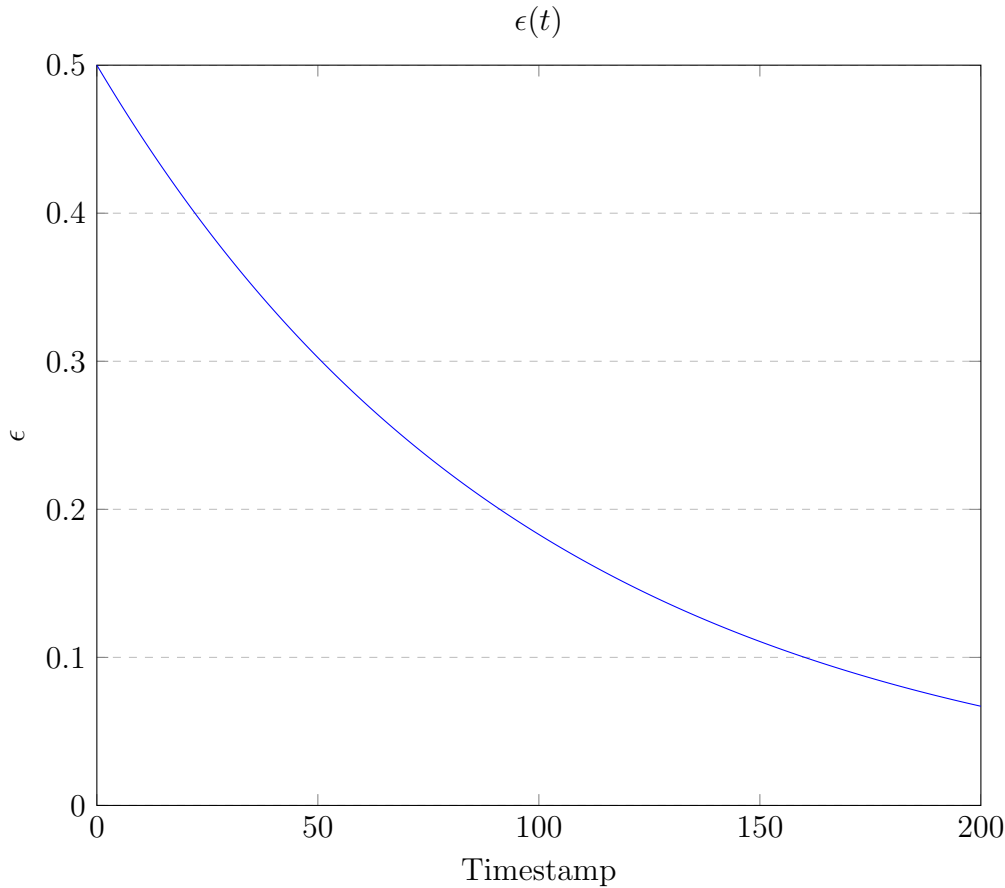

In doing so, the intervention values are randomly selected for exploitation with the most optimistic intervention value and exploration by mixing in alternative intervention values. 
It is typical for on-policy learning algorithms that exploration is more valuable in the beginning when the gathered data is little and the model of the environment is imprecise. Here the environment model, i.e., the PAG $\mathcal{G}$, is also imprecise initially, but with increasing data, it allows for better predictions. 
Therefore it makes sense to explore more initially and eventually exploit more as the interventions become more optimal. 
This shift towards more exploitation is controlled by decaying $\epsilon$-value. Initially $\epsilon= 0.5$, then each timestep $\epsilon$ is updated with $\epsilon = \epsilon*0.99$.
Figure \ref{fig:eps} shows the resulting epsilon function across time.

Now we learned how the tradeoff between taking the most optimistic intervention value and the alternative intervention value is handled. 
In the following, I first describe how to find the optimal intervention $do(X^{i_{opt}}=x_{opt})$ and then the alternative intervention value $x_{alt}$.

\subsection{Finding the optimal intervention value}
This section explains how we select an intervention with the goal of maximizing the value of a target variable. At first, I give an overview of the problem, then go into the details, and the pseudocode can be found in Algorithm \ref{alg:do_opt} 

\subsubsection{Overview}
 In Section \ref{sec:modLPCMCI}, we learned how the modified LPCMCI algorithm computes a PAG $\mathcal{G}$ with its effect sizes. 
Now we want to use this information to select an intervention to maximize the value of a target variable $X^y$.
The PAG $\mathcal{G}$ can represent many MAGs (described in Section \ref{sec:graph}).
We can use one MAG and its effect sizes to reconstruct the SCM $\mathcal{M}$ over the observed variables.
With the reconstructed SCM $\mathcal{M}$, we can simulate the outcomes of potential interventions and declare the best intervention with the best outcome as the optimal intervention given this MAG.
However, there is the problem that the initial PAG $\mathcal{G}$ of LPCMCI may represent many MAGs, which might lead to conflicting answers with respect to the optimal intervention. How should we resolve the conflict?
In short, we will compute the optimal intervention for each MAG and then choose the intervention that is most optimistic for the intervention proposal and ignore the other less optimistic interventions.

In the following, I will describe this process of finding the most optimistic intervention in more detail, starting with a more formal problem description.

\subsubsection{Probelm description}
The goal is to find the intervention $do(X^{i_{opt}}=x_{opt})$ that maximizes the value of a target variable $X^y$, given an SCM $\mathcal{M}$. Due to potential time lags, the effect of interventions can be delayed, and I am specifically interested in $\ATE$ (the average outcome of a target variable over a time horizon T), given an intervention $do(X^{i}= x)$.
The optimal intervention $do(X^{i_{opt}}=x_{opt})$ given a SCM $\mathcal{M}$ is determined by the optimization problem
\begin{equation}
\label{eq:argmax_in_R}
(X^{i_{opt}}, x_{opt})= \operatorname*{argmax}_{(X^{i}, x)} \frac{1}{T}\sum_t^T X^y_t(\mathcal{M}_{do(X^{i}= x)})  \quad \text { subject to } \quad  0 \leq i< N, \quad x \in \mathcal{R}
\end{equation} 
meaning we search for the intervention variable $X^{i}$ and intervention value $x$ given a SCM $\mathcal{M}$ such that $\mathcal{M}$ maximizes\footnote{User the operator $\operatorname*{argmin}_{(X^{i}, x)}$ instead of $\operatorname*{argmax}_{(X^{i}, x)}$ to minimize the target variable $X^y$.}
$\ATE$.
To solve the optimization problem, I rely on brute force evaluation in the numerical experiments, as it is reliable and suitable for many use cases. Yet, depending on the application, there might be optimization techniques that are better suited \cite{foulds_optimization_2012}.

There is one more problem to solve equation \ref{eq:argmax_in_R} through brute force evaluations. There is an infinite search space due to $x\in \mathcal{R}$.
To solve this problem, I limit the search space to a finite list of interventions $\textbf{L}_{do} = \{do(X^{i}= x_0), \dots \}$, which changes \ref{eq:argmax_in_R} to
\begin{equation}
\label{eq:argmax_in_S}
do(X^{i_{opt}}= x_{opt})= \operatorname*{argmax}_{do(X^{i}= x)\in \mathbf{S_{do}}} \frac{1}{T}\sum_t^T X^y_t(\mathcal{M}_{do(X^{i}= x)}).
\end{equation}

While the code underlying this thesis is limited to finding a solution to this equation, the code is modular, such that the optimization method can easily be adapted or replaced.
For example, if one prefers to optimize several variables, the researcher should use a multi-objective optimization technique instead.

To summarize, we search for an intervention $do(X^{i}= x)$ from a list of interventions $\mathbf{L_{do}}$ that leads to the best outcome averaged across a time horizon, given an SCM $\mathcal{M}$. The next section explains what to consider when choosing a list of interventions $\textbf{L}_{do}$.

\subsubsection{Choosing the list of interventions}
The interventions that $\textbf{L}_{do}$ should consist of are highly application specific and can be adapted by the researcher.
In the following, I first describe what one should consider when creating $\textbf{L}_{do}$ and then describe the $\textbf{L}_{do}$ used in my numerical experiments.

Generally, it is the case that if there are many interventions in $\textbf{L}_{do}$, it has the advantage that the approximation often becomes more accurate but has the downside of an increased computational load.
However, if there are interventions where we know beforehand that they can not be optimal, we can exclude them without sacrificing approximation accuracy.
In real-world applications, optimal intervention values are typically not infinite but are bounded within an interval $x\in [a,b]$, where $a>-\infty$ is the lowest possible value and $b<\infty$ is the largest possible value. As these bounds are application specific, they should be modified by the researcher.
Furthermore, the discretization frequency of the interval $[a,b]$ can also depend on the application. 

For example, if all mechanisms $f_i$ in the SCM $\mathcal{M}$ are linear, then the relationship between $x$ and $X^y$ is also linear for each intervention variable $X^i$ and the optimum lies on a boundary.
For non-linear $\mathcal{M}$, further evaluations between $a$ and $b$ may be needed. It is typically the case that the higher the non-linearity of $\mathcal{M}$, the higher the discretization frequency should be.

The set of variables that can be intervened on is generally also restricted. For example, some variables are too expensive, difficult, or even impossible to intervene on.
Generally, at least the target variable $X^y$ can not be intervened easily. Otherwise, one should not even use this algorithm and just intervene on the target variable directly with $do(X^y=x_{opt})$.
It is essential to exclude all variables from $\textbf{L}_{do}$ that can not be intervened on. Otherwise, the optimization may return an $X^{opt}$ that is practically not intervenable and, therefore, a useless result.
\\

In the following, I explain how the list of interventions $\textbf{L}_{do}$ is created in the numerical experiments of this thesis.
Choosing the interventional bounds $x\in[a,b]$ of variables usually requires background knowledge. To avoid relying on this background knowledge, I assume that all possible intervention values $x_i$ of a variable $X^i$ are bounded between the fifth and 95th percentile of that variable $X^i$.
Formally that is $x_i \in [P_{05}(X^i),P_{95}(X^i)]$. 
However, there is the problem that in the beginning, the measured time series $\textbf{X}^{i}$ has only a few samples, and there can be a large discrepancy between the percentiles of the measured distribution and the actual percentiles of the data generator.
To counteract this problem, a Gaussian-distribution \begin{equation}
\label{eq:gauss}
\mathcal{N}(\mu^{i_{}},\sigma^{i_{}})= \frac{1}{\sigma^{i_{}}\sqrt{2\pi}} 
  \exp\left( -\frac{1}{2}\left(\frac{X^{i_{}}_t-\mu^{i_{}}}{\sigma^{i_{}}}\right)^{\!2}\,\right)
  \end{equation}
  is fitted to the measured time series $\textbf{X}^{i_{}}$ with mean
\begin{equation}
{\mu^{i_{}}}=\frac{1}{T} \sum_{t=0}^{T-1} X^{i_{}}_t
\end{equation}
and variance
\begin{equation}
{\sigma^{2,i_{}}}=\frac{1}{T} \sum_{t=0}^{T-1}\left(X^{i_{}}_t-{\mu^{2,i_{}}}\right).
\end{equation}
Then the interventional bounds are the $5^{th}$ and $95^{th}$ percentiles of the Gaussian in Equation \ref{eq:gauss}: 
\begin{equation}
\label{eq:bounds_insightme}
  x^{i}\in\left[P_{05}\left(\mathcal{N}(\mu^{i_{}},\sigma^{i_{}})\right), P_{95}\left(\mathcal{N}(\mu^{i_{}},\sigma^{i_{}})\right)\right]. 
\end{equation}

Now that we have the boundaries of the interventional values for each variable, determining which intervention values from \ref{eq:bounds_insightme} we include to $\textbf{L}_{do}$ is simple. 
Since I assume that $\mathcal{M}$ of my application is linear, the optimal intervention values can only lie on a boundary. Therefore, values between the boundaries do not have to be evaluated.
Furthermore, in the numerical experiments, I assume all variables except the target variable can be intervened.
This results in a list of interventions $\textbf{L}_{do}$ with
\begin{equation}
    do(X^i = x^i) \in \textbf{L}_{do} \quad\text{for all}\quad   x^i \in \left\{P_{05}\left(\mathcal{N}(\mu^{i_{}},\sigma^{i_{}})\right), P_{95}\left(\mathcal{N}(\mu^{i_{}},\sigma^{i_{}})\right)\right\}
    \quad\text{and}\quad
    i\in \{0,...,N-1\}\backslash y.
\end{equation}

This concludes the explanation of how I create a list of interventions $\mathbf{L_{do}}$, in which we search for an intervention $do(X^{i}= x)$ that leads to the best outcomes, given an SCM $\mathcal{M}$. 
The following section explains how we obtain these SCMs that we since took for granted.

\subsubsection{Reconstructing SCM}
In Section \ref{sec:scm} Equation \ref{eq:scm} we definded SCM through the parents $pa(V^j)$, mechanisms $f_j$ and noise variables $\eta^j$ of all variables $V^j$ with $j=0,\dots ,\tilde{N-1}$.
Since the goal is to reconstruct SCMs to simulate the outcomes of potential interventions, the noise terms can be neglected with $\eta^j=0$.
Furthermore, we only reconstruct the SCM over the measured variables $X^j$ with $j=0,\dots,{N-1}$, as we do not have information about the latent variables and assume that unmeasured variables can not be intervened on.
With this in mind, we can reconstruct an unintervened linear causally stationary SCM $\mathcal{M}$ over the observed variables with
\begin{equation}
\label{eq:scm_reconstructed}
X_{t}^{j}:=\sum_{i} f_{i}\left(X_{t-\tau_{i}}^{i}\right) \quad \text { for } \quad t \in\{0, \ldots, T-1\}, \quad i,j \in\{0, \ldots, {N-1} \}.
\end{equation}
The only information that we need is the mechanisms $f_i$, which are in the linear case qualitatively given by a MAG represented by the PAG $\mathcal{G}$ and quantified by its effect sizes (described in Section \ref{sec:theory:effect_sizes}).

The following section explains how we use the reconstructed SCM to simulate the outcomes of potential interventions. 

\subsubsection{Simulating the outcomes of an intervention}
Simulating the outcomes of an intervention in the SCM of Equation \ref{eq:scm_reconstructed} makes use of Algorithm \ref{alg:data-generato}.
Please note the following changes in notation compared to the Algorithm. 
It was defined with $\tilde{N}$ and $\textbf{V}$, which means it also generates the latent variables. However, here, we only generate from the observed variables, and thus I denote the ${N}$ generated variables with $\textbf{X}^i$ instead of $\tilde{N}$ variables $\textbf{V}^i$.

To generate data, the SCM needs a starting sequence $\mathbf{X}_{start}$ of length $\tau_{max}$ if the maximal time lag is greater than zero (in Line \ref{ln:req} of Algorithm \ref{alg:data-generato}).
Future outcomes can be best simulated when the starting sequence $\mathbf{X}_{start}$ consists of the $\tau_{max}$ most recent measurements. In the numerical experiments of this thesis, this is always the case, meaning $\mathbf{X}_{start}$ consists of the most recent measurements of $\mathbf{X}$ from Algorithm \ref{alg:loop} Line \ref{ln:measurements}.

After initialization, the reconstructed SCM $\mathcal{M}$ is intervened (Section \ref{sec:do}) with with $do(X^{i}_t=x)$ yielding the intervened SCM $\mathcal{M}_{do(X^{i}_t=x)}$.
$\mathcal{M}_{do(X^{i}_t=x)}$ generates an intervened time-series $\textbf{X}$ of length $T$. Note that this intervention is conducted for every timestep $0 \leq t <T$.
From this simulated intervened time-series $\textbf{X}$, we now compute the average outcome $\ATE$ of one intervention.

In case the current initialization measurements are unavailable, one can compute the outcome $\ATE$ for several random initializations and average over these outcomes.

Now we have all the information to solve Equation \ref{eq:argmax_in_S} for a given SCM $\mathcal{M}$. 
However, the causal discovery Algorithm \ref{alg:modLPCMCI} does not return a MAG, but a whole equivalence class of MAGs, which might lead to different intervention proposals.
The following section describes which intervention proposals we listen to and which we ignore.

\subsubsection{Optimism in the face of uncertainty}

Algorithm \ref{alg:modLPCMCI} does not give a fine-grained believability estimation of the MAGs represented by the PAG $\mathcal{G}$.
It is only binary: If the MAG is not included by the PAG $\mathcal{G}$ we believe the MAG can not represent the data-generating process over the observed variables, and if it is represented by the PAG $\mathcal{G}$ it may represent the data-generating process.
This means we cannot just act according to the most believable MAGs, as all are equally believable models. 
However, the following example shows which graph we would listen to intuitively.
Consider two equally believable models.
The first recommends intervention A and predicts that it has little effect, while the second model recommends action B and predicts a large improvement. 
In the face of this uncertainty, it is intuitive to favor listening to the optimistic recommendation, even if it is not more believable.
The principle that is conveyed in this example is called \textit{optimism in the face of uncertainty} \cite[Section 34.4.5]{murphy_probabilistic_2023}. 

Like in the example, we search for the optimal intervention according to each MAG in the PAG $\mathcal{G}$ and only follow the intervention of the most optimistic MAG predicting the best outcome.
This means that the estimated outcome of the chosen intervention is optimistically biased. However, this will be corrected with more data.
If the selected intervention $do(X^i_t=x)$ repeatedly does not affect the target variable $X^y_t$, the intervention $do(X^i_t=x)$ will not be selected anymore because we learn through interventional discovery that $X^i_t \notin an(X^y_t)$ which removes the optimism bias of intervention $do(X^i_t=x)$.
If the selected intervention $do(X^i_t=x)$ affects the target variable $X^y_t$ and the effect sizes are correct, then the optimal intervention is found. If the effect sizes are wrong, they are improved through additional data.

Optimism in the face of uncertainty ensures that interventions are avoided that yield neither potentially valuable outcomes nor informative causal data.
Algorithm \ref{alg:do_opt} describes the implementation for this thesis.

\begin{algorithm}
\caption[Control algorithm]{Control algorithm to find the optimal univariate intervention given a PAG $\mathcal{G}$}\label{alg:do_opt}
\begin{algorithmic}[1]
\Require 
PAG $\mathcal{G}$ with effect sizes, list of all intervenable variables $X^i$ with a set $\textbf{S}^i_x$ of possible intervention values $\mathbf{L_{do(x)}}=\left[\left(X^i, \textbf{S}^i_x\right), \left(X^j, \textbf{S}^j_x\right),\dots\right]$, 
target variable $y$
\State  $O_{max} \gets 0$ \Comment{initialize the maximal effect variable}
\State  $X^{i_{opt}} \gets $ $None$ \Comment{initialize the optimal variable to intervene}
\For{each MAG in the Markov equivalence class represented by $\mathcal{G}$}
    \State translate the MAG with the effect sizes of $\mathcal{G}$ to a SCM $\mathcal{M}$
    \For{each variable $X^i$ in intervenable variables $\mathbf{L_{do(x)}}$}
        \For{each intervention value $x^i_k$ in the set of possible intervention values $\textbf{S}^i_x= \{x^i_{0}, x^i_{1}, \dots\}$}
            \State Let the SCM of $\mathcal{M}$ generate a time series $\mathbf{\bar{X}}$ with length $T$ with interventions $do\left(X^i_t=x^i_k\right)$, 
            \State $O \gets \frac{1}{T} \sum_{t}X^y_t$  \Comment{average outcome}   
            \If{$|O| > |O_{max}|$}
                \State $x_{opt} \gets x^i_k$
                \State $X^{i_{opt}} \gets X^{i_{}} $
            \EndIf
        \EndFor
    \EndFor
\EndFor
\State\Return optimal intervention $do(X^{i_{opt}}=x_{opt})$
\end{algorithmic}
\end{algorithm}

\subsection{The alternative intervention value}
There is an exploration-exploitation-tradeoff in choosing $x_{alt}$. In a linear setting, if the difference between the proposed treatment and the alternative treatment $\left(| x_{opt}-x_{alt}|\right)$ is large, then the outcomes of the alternative interventions are expected to be far from optimal, and we do not exploit the knowledge we already have. On the positive side, if $\left(| x_{opt}-x_{alt}|\right)$ is large, then differences in the outcomes between the treatments are also more considerable, leading to more robust independence tests during interventional discovery, which leads to better estimations of the causal model $\mathcal{G}$, allowing for potential better intervention in the future.
In the case of the actual application of this thesis, the users might tend to be unsatisfied as soon as the intervention leads to worse results than not intervening. Therefore I set the alternative intervention to the $50^{th}$ percentile (median). 




%

\section{Metrics and Evaluation}
\label{sec:average_regret}
The second research question of this thesis asks whether LPCMCI with interventional constraints can be used to learn a causal model, take the best action with respect to the model and improve the model with the resulting interventional data of the action.
It is furthermore interesting to quantify how the actions improve when additionally learning from interventional data.
To quantify how good actions are in a true SCM $(\mathcal{M}^*_l)$, I compute a unit called \textit{average regret} $\bar{R}$, which is the average of the differences between the outcomes of the actual actions taken ${X^{y}_t}$ and the theoretical outcomes if the optimal intervention would be taken every time ${X^{y}_t}^*$. Formally that is
\begin{equation}
\label{eq:average_regret}
    \bar{R_l}(\mathcal{M}^*_l) = \frac{1}{T}\sum^{T}_{t=1} \left({X^{y}_t}^*-X^{y}_t\right).
\end{equation}
The induced average regret $\bar{R_l}$ can vary depending on the randomly sampled SCM $(\mathcal{M}^*_l)$. Therefore each parameter setting is repeated 500 times, leading to a set of average regret $\left\{\bar{R}_1,\dots, \bar{R}_{500}\right\}$. 
To compare parameter settings with one metric, I again take the average of the set of average regret $\left\{\bar{R}_1,\dots, \bar{R}_{500}\right\}$ and with some abuse of notation denote it
\begin{equation}
\label{eq:average_average_regret}
    \bbR = \frac{1}{500}\sum^{500}_{l=1} \bar{R_l}.
\end{equation}
In the upcoming results chapter, this $\bbR$ is also shown as the mean line in Box-Plots.


\chapter{Results}
\label{ch:results}

This section shows the results of numerical experiments. If not explicitly adapted for an experiment, the parameters are set as described previously in Section \ref{ch:RQOne} and \ref{ch:RQTwo}.

\begin{figure}[htbp]
\centering
    \subfigure[The true data-generating SCM. The goal is to maximize the value of the target variable $X^0$ by intervening on any other variable. The best intervention would be to intervene on $X^2$ as it is the only variable affecting $X^0$. Although $X^6$ correlates even stronger with $X^0$, an intervention on $X^6$ will not help, as $X^6 \leftarrow X^0$.]{
         \includegraphics[width=0.6\linewidth]{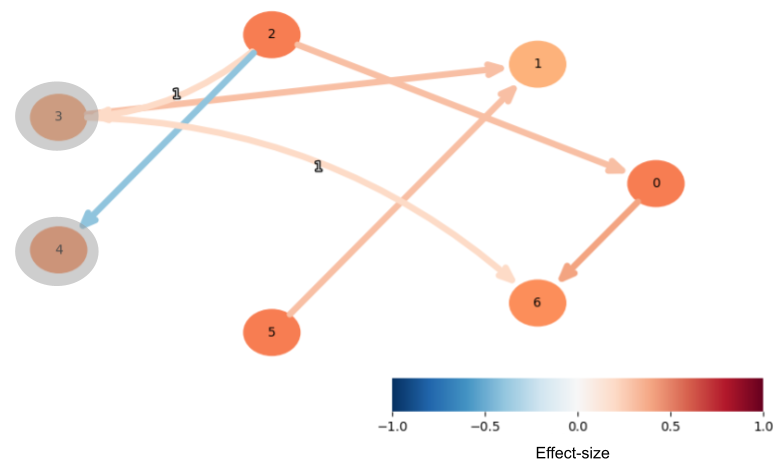}}\hspace{5mm}
             \subfigure[The causal discovery algorithm falsely writes $X^6 \rightarrow X^0$, which then leads to a useless intervention on $X^6$.]{
    \includegraphics[width=0.6\linewidth]{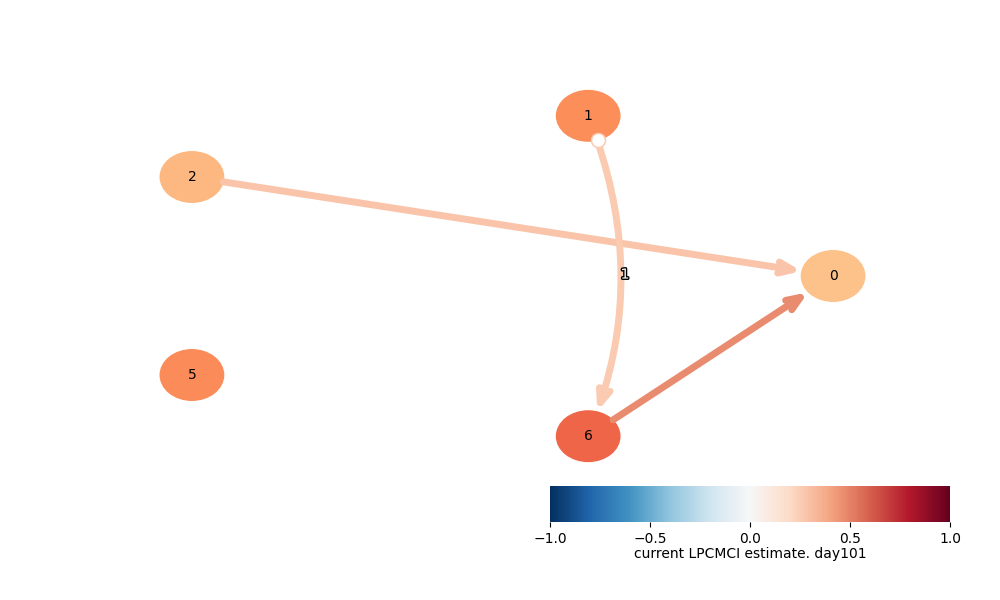}}\hspace{5mm}
         
     \subfigure[After ten useless interventions on $X^6$, the interventional causal discovery algorithm detects that $X^6 \notin an(X^0)$. This detection leads to to the correct edge orientation of $X^6 \leftarrow X^0$ resulting in a useful intervention on $X^0$.]{
         \includegraphics[width=0.6\linewidth]{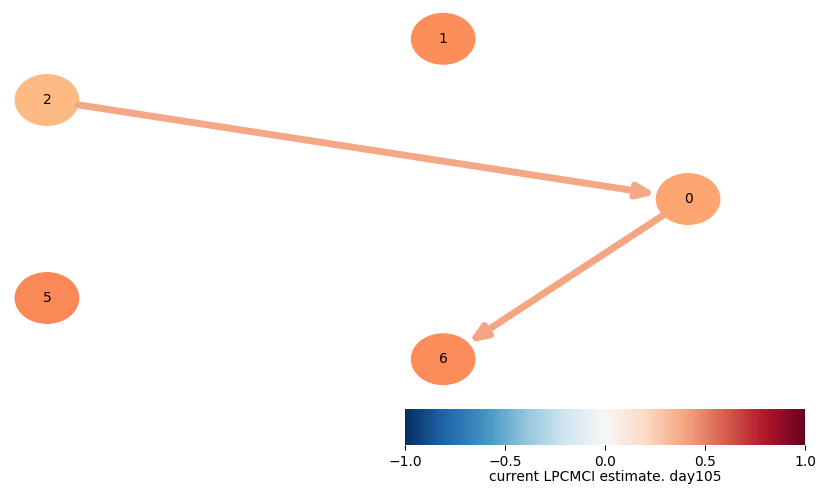}}\hspace{5mm}
    \caption{Results of numerical experiments}
        \label{fig:example-interventional-constraint}
\end{figure}

Figure \ref{fig:example-interventional-constraint} shows an example of a data-generating SCM. Nodes $X^3$ and $X^4$, which are partially occluded, represent two latent variables.
The goal is to maximize the value of a target variable $X^y$, here $y=0$, by intervening on any other variable. The best intervention would be to intervene on $X^2$ as it is the only variable affecting $X^0$. Although $X^6$ correlates even stronger with $X^0$, an intervention on $X^6$ will not help, as $X^6 \leftarrow X^0$.

My proposed algorithm intervenes $60.9\%$ of all interventions on the optimal variable $X^{opt}$, whereas the original LPCMCI algorithm without interventional constraints intervenes only to $53.6\%$ on $X^{opt}$. This is an increase of $7.3$ percent.
For comparison, when executing the numerical simulations but instead of interventional discovery, I use the ground truth as interventional constraints, then the optimal variable $X^{opt}$ is intervened $84.6\%$ of all interventions.
This is still imperfect, as the perfect score would be $100\%$. 
This means that observational causal discovery and the effect size estimation lead to an almost $16\%$ decrease.
This decrease can be explained as the observational causal discovery algorithm can remove true interventional constraints due to falsely detected conditional independencies in the observational data.
Therefore, the drop of $23.2\%$ from $84.1\%$ of optimal interventional constraints to $60.9\%$ of the proposed LPCMCI extension is probably due to imperfect and incomplete constraints from interventional causal discovery. 

\begin{figure}[htbp]
\centering
    \subfigure[The effect of the fraction of interventions on the average regret. There is a clear trend that average regret with more interventions.]{
         \includegraphics[width=0.45\linewidth]{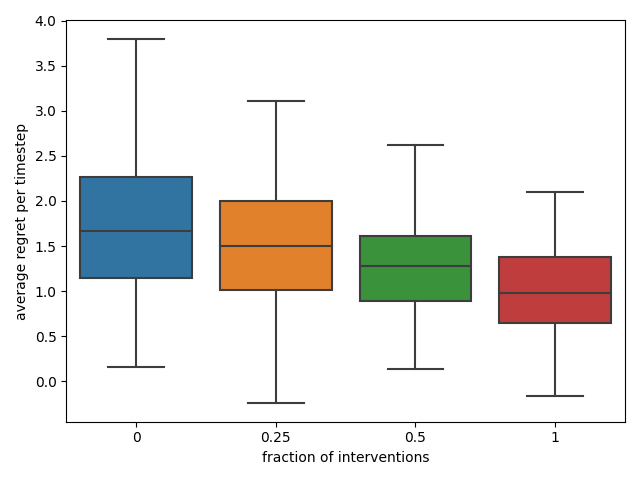}}\hspace{5mm}
             \subfigure[The extended LPCMCI vs. the original LPCMCI algorithm. The extended LPCMCI leads to slightly less regret than the original LPCMCI algorithm. ]{
    \includegraphics[width=0.45\linewidth]{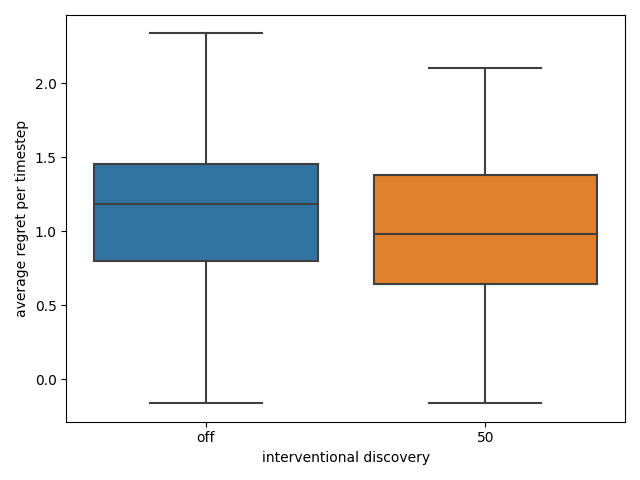}}
         
     \subfigure[The effect of the number of initial observations on average regret. An increase in initial observational data seems to have a positive impact.]{
         \includegraphics[width=0.45\linewidth]{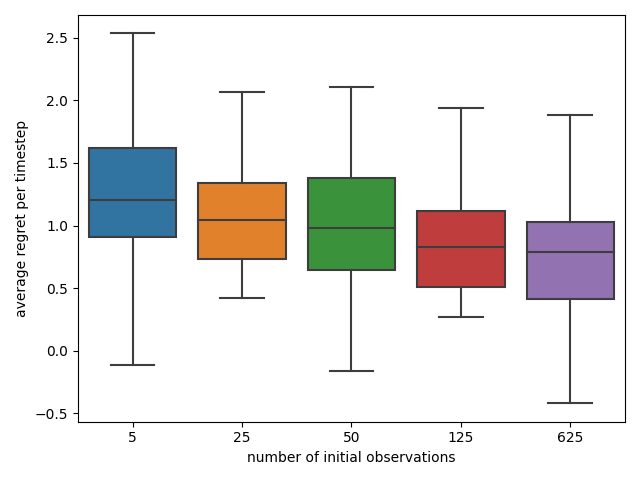}}\hspace{5mm}
     \subfigure[The effect of the number of observed variables on average regret. Here the number of latent variables is fixed at 2. This result indicates that an increase in the number of variables leads to more regret, except $N=8$. ]{
         \includegraphics[width=0.45\linewidth]{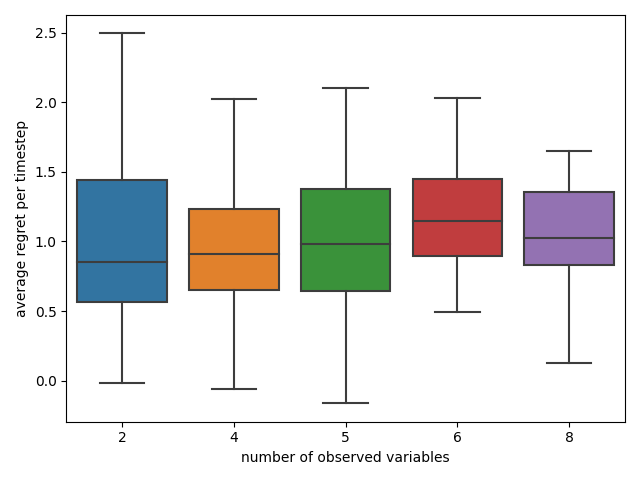}}
    \caption{Results of numerical experiments}
        \label{fig:results_numerical}
\end{figure}

\begin{figure}[htbp]
\centering
    \subfigure[The effect of $\alpha_{LPCMCI}$ on average regret. While $\alpha_{LPCMCI}=0.05$ leads to the lowest regret, it seems to have hardly any impact on regret.]{
        \includegraphics[width=0.45\linewidth]{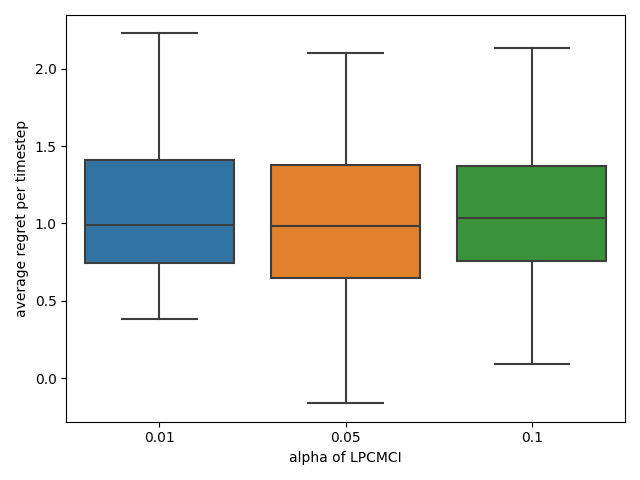}}\hspace{5mm}
    \subfigure[The effect of $\alpha_{\ind}$ on the average regret. $\alpha_{\ind}=0.8$ leads to the lowest average regret while larger and smaller values of $\alpha_{\ind}$ increase the average regret.]{
        \includegraphics[width=0.45\linewidth]{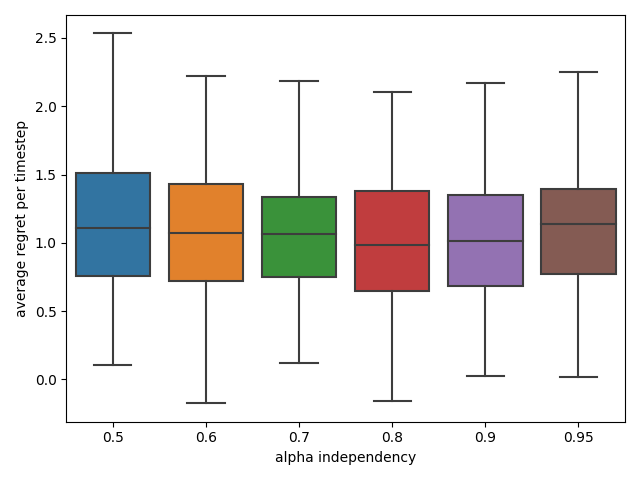}}
    \subfigure[The effect of $\alpha_{\dep}$ on the average regret. $\alpha_{\dep}=0.1$ leads to the lowest average regret, while larger and smaller values of $\alpha_{\dep}$ increase the average regret.]{
         \includegraphics[width=0.49\linewidth]{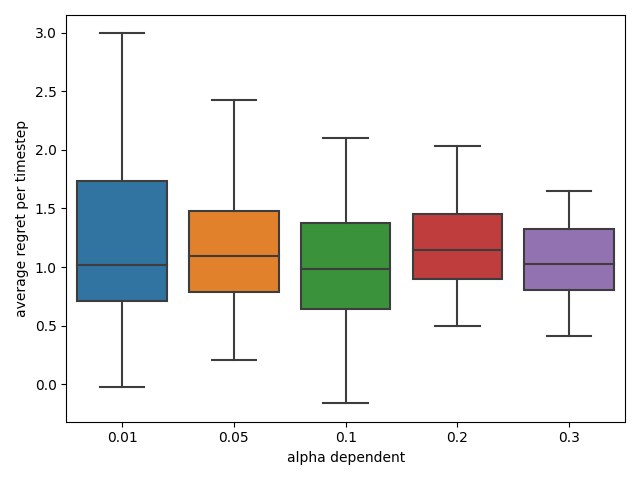}}\hspace{5mm}
    \caption{Results of numerical experiments. The effect of various variables on the average regret.}
        \label{fig:results_numerical2}
\end{figure}

Figure \ref{fig:results_numerical} and Figure \ref{fig:results_numerical2} show the effect of various variables on the average regret.
Figure \ref{fig:results_numerical} (a) shows the effect for different fractions of interventions. A fraction of zero interventions means that there are no interventions and only observational data exists.
The mean average regret $\bbR$ (explained in Section \ref{sec:average_regret}) decreases from $1.7$ when there are no interventions to $1.0$ when there is an intervention at each timestep.
This clearly shows that interventions recommended through the extended LPCMCI algorithm help reduce regret.

Figure \ref{fig:results_numerical} (b) compares the original LPCMCI that does not use interventional data and the extended LPCMCI algorithm with interventional discovery. The default setting with interventional discovery leads to an average regret $\bbR$ of 1.0, while the old LPCMCI algorithm leads to average regret $\bbR$ of 1.2.
This indicates that interventional discovery is useful, but the effect is, unfortunately, small. 
Explanations for this low difference are that the original LPCMCI algorithm does not need alternative interventions $do(X^{opt}=x_{alt})$ to explore. Instead, all interventions exploit with the most optimistic intervention $do(X^{opt}=x_{opt})$.
Furthermore, interventional discovery only increases the probability of intervening on the optimal variable $do(X^{opt})$ from $53.6\%$ to $60.9\%$.

Figure \ref{fig:results_numerical} (c) shows that the average regret decreases from $1.2$ with only $5$ initial observations to $0.8$ when there are $625$ initial observations. This trend was expected as more observational data should make the conditional independence test more robust.

Figure \ref{fig:results_numerical} (d) shows the effect of the number of observed variables on the average regret. Here the number of latent variables is fixed at 2. This result indicates that an increase in the number of variables leads to more regret, except $N=8$. This general trend of decreasing performance with an increasing number of variables is also observed in the original LPCMCI algorithm \cite{gerhardus_lpcmci_2020}.

Figure \ref{fig:results_numerical2} (a) shows that the significance level of $0.05$ for observational causal discovery $\alpha_{LPCMCI}$ leads to the lowest average regret, but the difference to the other tested values is negligible. 

Figure \ref{fig:results_numerical2} (b) shows the effect of $\alpha_{\ind}$ on the average regret. The means of the average regrets seem to form a parabola with the lowest value for $\alpha_{\ind}=0.8$, while larger and smaller values of $\alpha_{\ind}$ increase the average regret. 

Figure \ref{fig:results_numerical2} (c) shows the effect of $\alpha_{\dep}$ on the average regret. $\alpha_{\dep}=0.1$ leads to the lowest average regret, while larger and smaller values of $\alpha_{\dep}$ increase the average regret. However, in contrast to $\alpha_{\ind}$, there is no clear parabola visible.

\chapter{Discussion}
\label{ch:discussion}

\section{Limitations and future work}
The proposed approach to extend LPCMCI to profit from interventional data has limitations. I explain these limitations and point to possible further research in the following.

\subsection{Empirical validation}
The empirical validation through numerical experiments is limited and could be extended to include a full parameter study where all parameter combinations are simulated.
Furthermore, it would be interesting to quantify the performance of several sub-algorithms when used in isolation. For example, compare the performance of the original LPCMCI algorithm and the extended version with perfect interventional constraints and evaluate the precision and recall of lagged, contemporaneous, auto-dependent, and cross-dependent links.

\subsection{Finding the optimal intervention}
\subsubsection{Brute force optimization}
The algorithm to find the optimal intervention (Algorithm \ref{alg:do_opt}) evaluates each possible intervention in each possible MAG compatible with the PAG by simulation. This brute force optimization becomes computationally infeasible for large graphs, so I only evaluated graphs with a maximum of eight observable variables.
The computational load could be reduced by not simulating the outcome of interventions when it is obvious that an intervention will not be optimal. For example, the intervention on a variable with no path to the target variable. 
Furthermore, for all linear SCMs, the simulation could be replaced by computing the potential outcomes analytically.

\subsubsection{Univaiate interventions}
The developed algorithm of this thesis only supports univariate interventions, meaning that only one intervention per timestep can be proposed. However, it may be the case that even better outcomes can be reached when intervening on multiple variables simultaneously. 

\subsubsection{Optimizing a single target variable}
Similarly to univariate interventions, the code of this thesis only supports single objective optimization, where only one variable can be selected for optimization at a time. Nevertheless, taking interventions that optimize for several variables may be helpful.

\subsection{Inability to recover from falsely discovered interventional indepnendencies}
Suppose that variable $X^1$ is actually the best intervention to optimize $X^y$. However, due to an unlikely constellation of random noises, the interventional data indicates that $X^1 \ind X^y$ when intervened on $X^1$. Consequently, interventional discovery (Algorithm \ref{alg:intervDiscov}) declares that $X^1 \notin an(X^y)$ and therefore there will be no intervention on $X^1$. Since there is no intervention on $X^1$, we do not gather new interventional data that shows that $X^1 \dep X^y$.
In this case, it, unfortunately, means that such a false belief will never be corrected, and the optimal intervention will not be taken anymore.
This strong downside might explain that the best $\alpha_{\ind}$ keeps the number of false detection of independencies low at the cost of missing many true independencies (see Section \ref{sec:intervDisc}.
One way to mitigate this problem is by choosing the variable to intervene on probabilistically.
Another mitigation is to change the alternative intervention to also randomize over variables instead of only their intervention values. 

\subsection{Binary intervention costs}
In real applications, interventions usually come at a cost, where some interventions are more difficult or expensive than others.
It might be the case that some interventions are possible but so expensive that they should only be proposed if the effect is worth it.
Unfortunately, this algorithm does not support fine-grained cost assignments, and the researcher has only the binary choice to include or exclude variables from the set of intervenable variables and can define a range of possible intervention values.

\subsection{The limitation of hard interventions}
The implementation of this thesis only works with hard interventions, meaning the intervened variable $X$ is set to a specified value $x$, irrespective of the original influences of the SCM. In contrast, there are also soft interventions, which keep the original influences of the SCM and influence the conditional distribution \cite{frederick_eberhardt_causation_2007}, i.e., by adding an additional influence to the SCM. In the intended application of this thesis, hard interventions are more valuable. For example, it is easier to conduct a hard intervention by following a recommendation to walk 10 000 Steps instead of the soft intervention by following the recommendation to walk an additional 5000 steps than I would have otherwise, as the default value "otherwise" is not known yet. However, in other applications, it may be too difficult to change the underlying system through hard interventions, but it may be possible just to perturb it through soft interventions.

\subsection{Can conditional independence tests of LPCMCI become more robust when including certain interventional data?} 
\label{sec:interventionalDataInLPCMCI}
As mentioned in Section \ref{sec:modLPCMCI}, the modified LPCMCI algorithm (Algorithm \ref{alg:modLPCMCI}) uses only the observational data $\mathbf{\bar{X}}$ for conditional independence tests.
In the following, I explain how one can increase conditional independence tests' robustness by including some interventional data.

To recall from Section \ref{sec:lpcmci}, LPCMCI conducts conditional independence tests $CI(X, Y, S)$ between a pair of variables $X$ and $Y$ while conditioning on (a possibly empty) set $S$. If $X\ind Y |S$, the edge between $X-Y$ is removed.

If a true connection between two variables $X$ and $Y$ were $X \rightarrow Y$, data with interventions on $X$ would show that $X \dep Y |S$. As a result, the link in $X \rightarrow Y$ will not be removed, which is correct. 
However, the following example shows that including interventional data can lead to false conclusions.
Suppose the actual connection between them is $X \leftarrow Y$. Intervening on $X$ makes $X$ independent of $Y$, and LPCMCI would falsely remove the edge. This consequence means one should exclude data where the intervention was on the effect variable.
Furthermore, variables of the conditioning set may be intervened variables.

Suppose there is a fork $X\leftarrow Z \rightarrow Y$ and $Z\in S$. LPCMCI usually expects that $X \ind Y |S\Rightarrow$. An intervention on $Z$ would not change this relationship.
In contrast, suppose there is a collider $X\rightarrow Z \leftarrow Y$ and $Z\in S$. LPCMCI usually expects that $X \dep Y |S\Rightarrow$. However, an intervention breaks this dependency, and LPCMCI would not identify this collider. This means one should exclude data where the intervention was on a collider.

These results show that we could only include interventional data for conditional independence tests if we knew the causal structure in advance. In a sense, this is the case for LPCMCI due to its iterative nature, as explained in Section \ref{sec:lpcmci_iterative}. Since there are already predictions of causal relationships while conditional independencies are still being discovered, there might be an opportunity to make these later conditional independence tests more robust by including interventional data if there was no intervention on a collider or effect variable.

\subsection{Assumptions}
Lastly, I want to repeat that this algorithm assumes faithfulness, casual stationarity, acyclicity, and the absence of selected variables. Furthermore, the numerical experiments were only conducted with linear SCMs.

\section{Conclusion}
Reconstructing the causal relationships behind the phenomena we observe is a fundamental challenge in all areas of science.

The first part of this thesis explains how the observational causal discovery algorithm LPCMCI\cite{gerhardus_lpcmci_2020} can be extended to profit from casual constraints found through randomized experiments.
I find that LPCMCI can be extended to use interventional data. My numerical results show that, given perfect interventional constraints, the reconstructed SCMs of the extended LPCMCI allow $84.6\%$ of the time for the optimal prediction of the target variable. This contrasts the baseline of $53.6\%$ when using the original LPCMCI algorithm that does not use interventional constraints. 
I could not achieve $100\%$ because interventional constraints describing dependencies can be lost during observational causal discovery when dependencies are not visible in the observational data.

The second part of this thesis investigates the question of regret minimizing control by simultaneously learning a causal model and planning actions through the causal model.
The idea is that an agent with the objective of optimizing a measured variable first learns the system's mechanics through observational causal discovery. The agent then intervenes on the most promising variable with randomized values allowing for the exploitation and generation of new interventional data.
In contrast to the first part, the true interventional constraints are not given but must be detected in the interventional data. Then these interventional constraints are used to enhance the causal model further, allowing improved actions the next time that induce less regret.
I find that the extended LPCMCI can be favorable compared to the original LPCMCI algorithm. 
My numerical results show that detecting and using interventional constraints leads to reconstructed SCMs that allow $60.9\%$ of the time for the optimal prediction of the target variable. This result is in contrast to the baseline of $53.6\%$ when using the original LPCMCI algorithm and $84.6\%$ when using the true interventional constraints. The result also indicates that further performance improvements are possible with better interventional discovery.
Furthermore, the induced average regret decreases from $1.2$ when using the original LPCMCI algorithm to $1.0$ when using the extended LPCMCI algorithm with interventional discovery.
A disadvantage of the proposed regret-minimizing algorithm is the need for explorative interventions, which are not optimized for regret reduction. In contrast, when ignoring interventional discovery, explorative interventions are not needed.

I hope this thesis inspires future work, and its source code is available \href{https://github.com/christianreiser/correlate/blob/master/causal_discovery/LPCMCI/plan.py}{online}.

\appendix


\backmatter
\addcontentsline{toc}{chapter}{Bibliography}
\bibliographystyle{IEEEtranN}
\bibliography{./src/bib/references}

\begin{thebibliography}{28}
\providecommand{\natexlab}[1]{#1}
\providecommand{\url}[1]{#1}
\csname url@samestyle\endcsname
\providecommand{\newblock}{\relax}
\providecommand{\bibinfo}[2]{#2}
\providecommand{\BIBentrySTDinterwordspacing}{\spaceskip=0pt\relax}
\providecommand{\BIBentryALTinterwordstretchfactor}{4}
\providecommand{\BIBentryALTinterwordspacing}{\spaceskip=\fontdimen2\font plus
\BIBentryALTinterwordstretchfactor\fontdimen3\font minus
  \fontdimen4\font\relax}
\providecommand{\BIBforeignlanguage}[2]{{%
\expandafter\ifx\csname l@#1\endcsname\relax
\typeout{** WARNING: IEEEtranN.bst: No hyphenation pattern has been}%
\typeout{** loaded for the language `#1'. Using the pattern for}%
\typeout{** the default language instead.}%
\else
\language=\csname l@#1\endcsname
\fi
#2}}
\providecommand{\BIBdecl}{\relax}
\BIBdecl

\bibitem[Gerhardus and Runge(2020)]{gerhardus_lpcmci_2020}
A.~Gerhardus and J.~Runge, ``\BIBforeignlanguage{en}{{LPCMCI} - {High}-recall
  causal discovery for autocorrelated time series with latent confounders},''
  \emph{\BIBforeignlanguage{en}{arXiv}}, p.~11, Jul. 2020.

\bibitem[Durieu et~al.(2001)Durieu, Marguery, Giordano-Labadie, Journe, Loche,
  and Bazex]{durieu_photoaggravated_2001}
C.~Durieu, M.~C. Marguery, F.~Giordano-Labadie, F.~Journe, F.~Loche, and
  J.~Bazex, ``\BIBforeignlanguage{fre}{[{Photoaggravated} contact allergy and
  contact photoallergy caused by ketoprofen: 19 cases]},''
  \emph{\BIBforeignlanguage{fre}{Annales de dermatologie et de venereologie}},
  vol. 128, no. 10 Pt 1, pp. 1020--1024, Oct. 2001.

\bibitem[Peters et~al.(2018)Peters, Janzig, and
  Schölkopf]{peters_elements_2018}
\BIBentryALTinterwordspacing
J.~Peters, D.~Janzig, and B.~Schölkopf, \emph{\BIBforeignlanguage{en}{Elements
  of causal inference: foundations and learning algorithms}}.\hskip 1em plus
  0.5em minus 0.4em\relax The MIT Press, Nov. 2018, vol.~88. [Online].
  Available:
  \url{https://www.tandfonline.com/doi/full/10.1080/00949655.2018.1505197}
\BIBentrySTDinterwordspacing

\bibitem[Reiser(2022{\natexlab{a}})]{reiser_causal_2022}
C.~Reiser, ``\BIBforeignlanguage{en}{Causal discovery for time series with
  latent confounders},'' \emph{\BIBforeignlanguage{en}{Unpublished}}, p.~10,
  Apr. 2022.

\bibitem[Hariton and Locascio(2018)]{hariton_randomised_2018}
\BIBentryALTinterwordspacing
E.~Hariton and J.~J. Locascio, ``Randomised controlled trials—the gold
  standard for effectiveness research,'' \emph{BJOG : an international journal
  of obstetrics and gynaecology}, vol. 125, no.~13, p. 1716, Dec. 2018.
  [Online]. Available:
  \url{https://www.ncbi.nlm.nih.gov/pmc/articles/PMC6235704/}
\BIBentrySTDinterwordspacing

\bibitem[Gerhardus(2021)]{gerhardus_characterization_2021}
\BIBentryALTinterwordspacing
A.~Gerhardus, ``Characterization of causal ancestral graphs for time series
  with latent confounders,'' \emph{arXiv:2112.08417 [cs, stat]}, Dec. 2021,
  arXiv: 2112.08417. [Online]. Available: \url{http://arxiv.org/abs/2112.08417}
\BIBentrySTDinterwordspacing

\bibitem[Zhang(2008{\natexlab{a}})]{zhang_completeness_2008}
\BIBentryALTinterwordspacing
J.~Zhang, ``\BIBforeignlanguage{en}{On the completeness of orientation rules
  for causal discovery in the presence of latent confounders and selection
  bias},'' \emph{\BIBforeignlanguage{en}{Artificial Intelligence}}, vol. 172,
  no. 16-17, pp. 1873--1896, Nov. 2008. [Online]. Available:
  \url{https://linkinghub.elsevier.com/retrieve/pii/S0004370208001008}
\BIBentrySTDinterwordspacing

\bibitem[Tian(2012)]{tian_generating_2012}
\BIBentryALTinterwordspacing
J.~Tian, ``Generating {Markov} {Equivalent} {Maximal} {Ancestral} {Graphs} by
  {Single} {Edge} {Replacement},'' Jul. 2012, arXiv:1207.1428 [cs, stat].
  [Online]. Available: \url{http://arxiv.org/abs/1207.1428}
\BIBentrySTDinterwordspacing

\bibitem[Zhang(2008{\natexlab{b}})]{zhang_causal_2008}
J.~Zhang, ``\BIBforeignlanguage{en}{Causal {Reasoning} with {Ancestral}
  {Graphs}},'' \emph{\BIBforeignlanguage{en}{Journal of Machine Learning
  Research}}, p.~38, 2008.

\bibitem[Richardson and Spirtes(2002)]{richardson_ancestral_2002}
\BIBentryALTinterwordspacing
T.~Richardson and P.~Spirtes, ``\BIBforeignlanguage{en}{Ancestral graph
  {Markov} models},'' \emph{\BIBforeignlanguage{en}{The Annals of Statistics}},
  vol.~30, no.~4, Aug. 2002. [Online]. Available:
  \url{https://projecteuclid.org/journals/annals-of-statistics/volume-30/issue-4/Ancestral-graph-Markov-models/10.1214/aos/1031689015.full}
\BIBentrySTDinterwordspacing

\bibitem[Pearl(2000)]{pearl_causality_2000}
J.~Pearl, \emph{Causality: models, reasoning, and inference}.\hskip 1em plus
  0.5em minus 0.4em\relax Cambridge, U.K. ; New York: Cambridge University
  Press, 2000.

\bibitem[Bejos et~al.(2020)Bejos, Morales, and Sucar]{bejos_maximal_2020}
S.~Bejos, E.~Morales, and E.~Sucar, ``\BIBforeignlanguage{en}{Maximal
  {Ancestral} {Graphs} - {Part} 1: {Fundamentals}},'' 2020.

\bibitem[Runge(2018)]{runge_causal_2018}
\BIBentryALTinterwordspacing
J.~Runge, ``Causal network reconstruction from time series: {From} theoretical
  assumptions to practical estimation,'' \emph{Chaos: An Interdisciplinary
  Journal of Nonlinear Science}, vol.~28, no.~7, p. 075310, Jul. 2018,
  publisher: American Institute of Physics. [Online]. Available:
  \url{https://aip.scitation.org/doi/full/10.1063/1.5025050}
\BIBentrySTDinterwordspacing

\bibitem[Xiao et~al.(2008)Xiao, Dellandrea, Dou, and Chen]{xiao_what_2008}
Z.~Xiao, E.~Dellandrea, W.~Dou, and L.~Chen, ``What is the best segment
  duration for music mood analysis ?'' in \emph{2008 {International} {Workshop}
  on {Content}-{Based} {Multimedia} {Indexing}}, Jun. 2008, pp. 17--24, iSSN:
  1949-3991.

\bibitem[Peters et~al.(2013)Peters, Janzing, and
  Schölkopf]{peters_causal_2013}
\BIBentryALTinterwordspacing
J.~Peters, D.~Janzing, and B.~Schölkopf, ``Causal {Inference} on {Time}
  {Series} using {Restricted} {Structural} {Equation} {Models},'' in
  \emph{Advances in {Neural} {Information} {Processing} {Systems}},
  vol.~26.\hskip 1em plus 0.5em minus 0.4em\relax Curran Associates, Inc.,
  2013. [Online]. Available:
  \url{https://proceedings.neurips.cc/paper/2013/hash/47d1e990583c9c67424d369f3414728e-Abstract.html}
\BIBentrySTDinterwordspacing

\bibitem[Runge et~al.(2019)Runge, Nowack, Kretschmer, Flaxman, and
  Sejdinovic]{runge_pcmci_2019}
\BIBentryALTinterwordspacing
J.~Runge, P.~Nowack, M.~Kretschmer, S.~Flaxman, and D.~Sejdinovic, ``{PCMCI} -
  {Detecting} and quantifying causal associations in large nonlinear time
  series datasets,'' \emph{Science Advances}, vol.~5, no.~11, p.~15, 2019,
  publisher: American Association for the Advancement of Science. [Online].
  Available: \url{https://www.science.org/doi/10.1126/sciadv.aau4996}
\BIBentrySTDinterwordspacing

\bibitem[Pearl(2012)]{pearl_-calculus_2012}
J.~Pearl, ``The do-calculus revisited,'' \emph{arXiv preprint arXiv:1210.4852},
  2012.

\bibitem[Eberhardt and Scheines(2007)]{eberhardt_interventions_2007}
\BIBentryALTinterwordspacing
F.~Eberhardt and R.~Scheines, ``\BIBforeignlanguage{en}{Interventions and
  {Causal} {Inference}},'' \emph{\BIBforeignlanguage{en}{Philosophy of
  Science}}, vol.~74, no.~5, pp. 981--995, Dec. 2007. [Online]. Available:
  \url{https://www.cambridge.org/core/product/identifier/S0031824800005651/type/journal_article}
\BIBentrySTDinterwordspacing

\bibitem[Jaber et~al.(2020)Jaber, Kocaoglu, Shanmugam, and
  Bareinboim]{jaber_causal_2020}
\BIBentryALTinterwordspacing
A.~Jaber, M.~Kocaoglu, K.~Shanmugam, and E.~Bareinboim, ``Causal {Discovery}
  from {Soft} {Interventions} with {Unknown} {Targets}: {Characterization} and
  {Learning},'' in \emph{Advances in {Neural} {Information} {Processing}
  {Systems}}, vol.~33.\hskip 1em plus 0.5em minus 0.4em\relax Curran
  Associates, Inc., 2020, pp. 9551--9561. [Online]. Available:
  \url{https://proceedings.neurips.cc/paper/2020/hash/6cd9313ed34ef58bad3fdd504355e72c-Abstract.html}
\BIBentrySTDinterwordspacing

\bibitem[David and Khandhar(2022)]{david_double-blind_2022}
\BIBentryALTinterwordspacing
S.~David and P.~B. Khandhar, ``\BIBforeignlanguage{eng}{Double-{Blind}
  {Study}},'' in \emph{\BIBforeignlanguage{eng}{{StatPearls}}}.\hskip 1em plus
  0.5em minus 0.4em\relax Treasure Island (FL): StatPearls Publishing, 2022.
  [Online]. Available: \url{http://www.ncbi.nlm.nih.gov/books/NBK546641/}
\BIBentrySTDinterwordspacing

\bibitem[Johnson(1999)]{johnson_insignificance_1999}
\BIBentryALTinterwordspacing
D.~H. Johnson, ``The {Insignificance} of {Statistical} {Significance}
  {Testing},'' \emph{The Journal of Wildlife Management}, vol.~63, no.~3, pp.
  763--772, 1999, publisher: [Wiley, Wildlife Society]. [Online]. Available:
  \url{https://www.jstor.org/stable/3802789}
\BIBentrySTDinterwordspacing

\bibitem[Rouder et~al.(2009)Rouder, Speckman, Sun, Morey, and
  Iverson]{rouder_bayesian_2009}
\BIBentryALTinterwordspacing
J.~N. Rouder, P.~L. Speckman, D.~Sun, R.~D. Morey, and G.~Iverson,
  ``\BIBforeignlanguage{en}{Bayesian t tests for accepting and rejecting the
  null hypothesis},'' \emph{\BIBforeignlanguage{en}{Psychonomic Bulletin \&
  Review}}, vol.~16, no.~2, pp. 225--237, Apr. 2009. [Online]. Available:
  \url{http://link.springer.com/10.3758/PBR.16.2.225}
\BIBentrySTDinterwordspacing

\bibitem[Kruschke(2011)]{kruschke_introduction_2011}
\BIBentryALTinterwordspacing
J.~K. Kruschke, ``\BIBforeignlanguage{en}{Introduction to {Special} {Section}
  on {Bayesian} {Data} {Analysis}},''
  \emph{\BIBforeignlanguage{en}{Perspectives on Psychological Science}},
  vol.~6, no.~3, pp. 272--273, May 2011. [Online]. Available:
  \url{http://journals.sagepub.com/doi/10.1177/1745691611406926}
\BIBentrySTDinterwordspacing

\bibitem[Brydges and Gaeta(2019)]{brydges_analysis_2019}
\BIBentryALTinterwordspacing
C.~R. Brydges and L.~Gaeta, ``An {Analysis} of {Nonsignificant} {Results} in
  {Audiology} {Using} {Bayes} {Factors},'' \emph{Journal of Speech, Language,
  and Hearing Research}, vol.~62, no.~12, pp. 4544--4553, Dec. 2019, publisher:
  American Speech-Language-Hearing Association. [Online]. Available:
  \url{https://pubs.asha.org/doi/abs/10.1044/2019_JSLHR-H-19-0182}
\BIBentrySTDinterwordspacing

\bibitem[Reiser(2022{\natexlab{b}})]{reiser_predicting_2022}
\BIBentryALTinterwordspacing
C.~Reiser, ``Predicting and {Visualizing} {Daily} {Mood} of {People} {Using}
  {Tracking} {Data} of {Consumer} {Devices} and {Services},''
  \emph{arXiv:2202.03721 [cs, stat]}, Feb. 2022, arXiv: 2202.03721. [Online].
  Available: \url{http://arxiv.org/abs/2202.03721}
\BIBentrySTDinterwordspacing

\bibitem[Foulds(2012)]{foulds_optimization_2012}
L.~R. Foulds, \emph{\BIBforeignlanguage{en}{Optimization {Techniques}: {An}
  {Introduction}}}.\hskip 1em plus 0.5em minus 0.4em\relax Springer Science \&
  Business Media, Dec. 2012, google-Books-ID: lybvBwAAQBAJ.

\bibitem[Murphy(2023)]{murphy_probabilistic_2023}
\BIBentryALTinterwordspacing
K.~P. Murphy, ``Probabilistic {Machine} {Learning} {Advanced} {Topics} {Kevin}
  {P}. {Murphy}.pdf,'' 2023. [Online]. Available: \url{probml.ai}
\BIBentrySTDinterwordspacing

\bibitem[Eberhardt(2007)]{frederick_eberhardt_causation_2007}
\BIBentryALTinterwordspacing
F.~Eberhardt, ``Causation and {Intervention},'' Ph.D. dissertation, Carnegie
  Mellon University, 2007. [Online]. Available:
  \url{https://www.its.caltech.edu/~fehardt/papers/PhDthesis.pdf}
\BIBentrySTDinterwordspacing

\end{thebibliography}

\end{document}